\documentclass[letterpaper]{article}
\usepackage{aaai2027}
\usepackage[hyphens]{url}
\usepackage{graphicx}
\usepackage{natbib}
\usepackage{caption}
\usepackage{amsmath}
\usepackage{amssymb}
\usepackage{booktabs}
\usepackage{algorithm}
\usepackage{algorithmic}
\usepackage{float}

\newcommand{\method}{Adaptive-WAM}

\title{Adaptive-WAM: Quality-Guided Early-Exit Planning\\from Intermediate Video-Diffusion Features}
\author{
Sining Ang\textsuperscript{\rm 1,2},
Yuguang Yang\textsuperscript{\rm 1,3},
Yan Wang\textsuperscript{\rm 1}\corresponding
}
\affiliations{
\textsuperscript{\rm 1}Institute for AI Industry Research (AIR), Tsinghua University\\
\textsuperscript{\rm 2}Department of Automation, University of Science and Technology of China\\
\textsuperscript{\rm 3}School of Electronic Information Engineering, Beihang University\\
angsn@mail.ustc.edu.cn, wangyan@air.tsinghua.edu.cn
}

\begin{document}

\nocopyright
\maketitle

\begin{abstract}
Large video diffusion models provide rich spatiotemporal priors for autonomous driving, but existing world-action models often inherit the cost of iterative future-video generation even though deployment only requires an ego trajectory.
We ask a more basic question: how much of a video diffusion model must be executed to make a reliable driving decision?
Through a controlled study of video denoising timesteps and Diffusion Transformer (DiT) depth, we find that planning performance is largely insensitive to the tested video-noise levels, whereas strong trajectories can already be decoded from intermediate layers.
Based on this observation, we introduce \method, a quality-aware multi-exit planner built on a Wan2.2-5B backbone.
Trajectory diffusion heads are attached to selected DiT blocks, and a lightweight trajectory-quality scorer terminates inference once the best trajectory decoded so far satisfies a quality threshold; otherwise, computation continues from the cached hidden state to a deeper exit.
The deployed planner therefore avoids the iterative classifier-free denoising loop and VAE decoding required for future-video synthesis, while dynamically allocating backbone depth according to trajectory quality.
On NAVSIM, the adaptive single-trajectory planner achieves 90.8 PDMS; a
separate fixed-exit variant reaches 92.6 PDMS with 64 proposals.
It further obtains 89.9 EPDMS on NAVSIM v2, yielding the best reported results among the compared front-view video world-model planners.
Without target-domain fine-tuning, \method\ transfers to nuScenes with 0.88\,m average L2 error and a 0.08\% collision rate.
On an A100, adaptive routing improves PDMS from 90.62 to 90.79 while averaging
170\,ms end-to-end planning latency, approximately 10\% below the 190\,ms
fixed block-15 planner and 47\% below the 320\,ms fixed full-depth planner.
Code will be released.
\end{abstract}

\section{Introduction}

Planning in open-world traffic requires more than recognizing the current
scene: an autonomous policy must anticipate how surrounding actors and road
context may evolve under its decisions.
Large video models provide temporal and motion priors for this purpose, and
driving-specific world models adapt such priors to controllable future
prediction, simulation, and planning
\cite{wan2025,hu2023gaia,gao2023magicdrive,gao2024vista}.
Video generation is therefore an appealing representation-learning objective
for autonomous driving.

Recent world-action models (WAMs) connect predictive representations to
control either by decoding actions from video-model features or by jointly
generating visual futures and trajectories
\cite{drivelaw2026,driveva2026,drivinggpt2025,fastwam2026}.
Although these designs strengthen the connection between prediction and
action, their deployed computation remains either coupled to iterative future
rollout or tied to a predetermined backbone path.
This is unnecessarily rigid: deployment requires only a low-dimensional ego
trajectory, and the quality of a plan may become sufficient before the full
video backbone has been evaluated.

\begin{figure*}[t]
    \centering
    \includegraphics[width=0.94\textwidth]{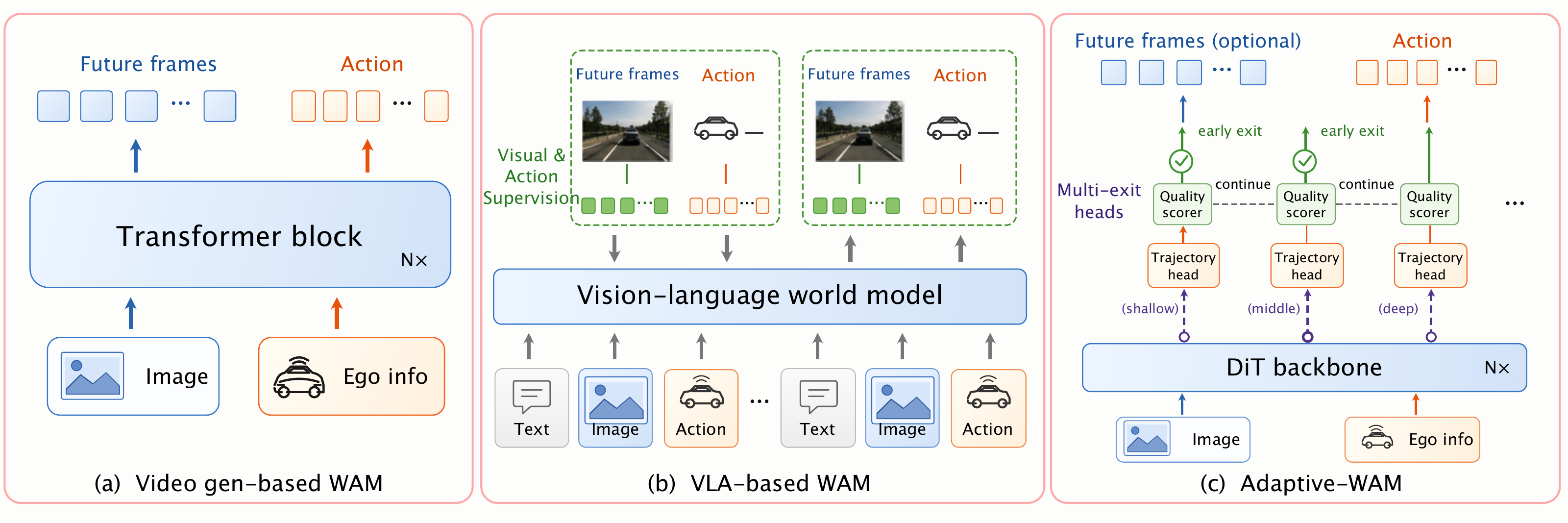}
    \caption{Comparison of predetermined and adaptive WAM interfaces.
    (a) Video-backbone WAMs follow a predetermined frame/action generation
    path; (b) multimodal WAMs generate visual and action streams along a
    predetermined path; (c) \method\ decodes one trajectory per attempted exit
    and routes by predicted quality. Future-video prediction supervises
    training but is not required by the deployed planner.}
    \label{fig:paradigm_comparison}
\end{figure*}

We therefore ask: how much of a video diffusion model must be executed to make
a reliable driving decision?
We separate two axes that are often conflated: the \emph{video diffusion
timestep}, which controls the latent noise level, and the \emph{DiT depth},
which controls how many transformer blocks are evaluated.
On NAVSIM \cite{navsim2024}, five tested video timesteps change the score of a
fixed layer by at most 0.15 points, whereas fixed exits exhibit substantial
depth-dependent variation in action quality.
The best intermediate exit outperforms the full-depth exit, yet the final
adaptive policy exceeds every fixed single-trajectory exit.
This evidence motivates allocating computation according to the quality of the
current plan rather than committing to one readout depth.

Based on this observation, we propose \method, a quality-aware, layer-adaptive world-action model.
We retain a Wan2.2-TI2V-5B \cite{wan2025} backbone and attach a ReCogDrive-style five-step trajectory DiT \cite{recogdrive2025} to six intermediate blocks.
Adaptive routing decodes one trajectory at each attempted exit and retains the
best trajectory accumulated across the evaluated exits.
A lightweight DINOv2-Small \cite{dinov2} scorer predicts NAVSIM planning
sub-scores for each decoded trajectory.
If the highest predicted score among the attempted exits passes a threshold,
the controller returns its trajectory; otherwise, backbone execution continues
to the next exit.
The default planning path skips the remaining video denoising loop, the unconditional classifier-free-guidance branch, and VAE video decoding.
Because trajectory rewards are highly saturated and frequently tied, the scorer
predicts metric components and acts primarily as an exit-quality verifier
rather than imposing a strict total ordering over trajectories.
We therefore evaluate both tie-aware selection and consequential large-gap
errors.
Across scorer-backbone ablations with Wan, ResNet, ViT, and DINO features, the
best Wan variant improves the diagnostic score by only 0.03 points over
DINOv2-Small while incurring substantially higher inference cost.

Experiments support three conclusions.
First, intermediate diffusion features provide a strong planning substrate:
the best fixed-exit model improves from 86.56 PDMS after imitation learning to
90.62 after DiffGRPO-style refinement, while the adaptive single-trajectory
planner reaches 90.79 PDMS.
Second, adaptive exits improve the performance--computation frontier.
At the selected threshold, the adaptive policy reaches 90.79 PDMS versus
90.62 for the strongest fixed single-trajectory exit.
Its average end-to-end planning latency is 170\,ms, approximately 10\% lower
than the 190\,ms fixed block-15 planner and 47\% lower than the 320\,ms fixed
full-depth planner.
Third, the learned representation transfers across datasets, reaching 0.88\,m average L2 error and 0.08\% collision rate on nuScenes \cite{nuscenes2020} without target-domain fine-tuning.

Our contributions are:
\begin{itemize}
    \item We systematically diagnose how video-noise level and DiT depth affect driving, revealing robustness to the five tested video-noise levels and distinct depth-dependent quality--computation trade-offs.
    \item We introduce a multi-exit world-action architecture whose learned trajectory-quality controller dynamically selects the required DiT depth without completing video generation.
    \item We evaluate planning, transfer, trajectory scoring, and efficiency on NAVSIM v1/v2 and nuScenes, obtaining state-of-the-art planning performance among compared world-model methods while improving over every fixed single-trajectory exit at lower average cost than the strongest one.
\end{itemize}

\section{Related Work}

\paragraph{Driving video generation and simulation.}
Driving world models learn controllable future observations from video.
GAIA-1 autoregressively models discrete visual tokens
\cite{hu2023gaia}; MagicDrive, DriveDreamer, Panacea, and DrivingDiffusion
condition diffusion models on structured geometry, scene layouts, or driving
controls
\cite{gao2023magicdrive,wang2024drivedreamer,wen2024panacea,
li2024drivingdiffusion}.
VISTA and MiLA further improve controllability and long-horizon consistency
\cite{gao2024vista,wang2025mila}, whereas ReSim and OmniNWM use generative
world models for controllable simulation and interactive data generation
\cite{yang2025resim,li2025omninwm}.
These studies establish video prediction as a source of spatiotemporal priors.
We investigate how much of this trained representation must be evaluated for
planning.

\paragraph{World-action models for planning.}
Latent world models use predicted future representations as planning
supervision or context.
LAW learns a latent dynamics model for end-to-end driving, DrivingWorld and
DriveWorld pretrain video-centric representations, and PWM jointly predicts
future states and actions
\cite{li2024enhancing,hu2024drivingworld,min2024driveworld,
zhao2025forecasting}.
More recent WAMs directly couple visual prediction and control.
DrivingGPT, Epona, VaViM/VaVAM, DriveVLA-W0, and FutureSightDrive model future
observations and actions through autoregressive, parallel, or intermediate
reasoning streams
\cite{drivinggpt2025,epona2025,vavim2025,drivevlaw02025,fsdrive2025}.
DriveLaW conditions an action diffuser on video-model features, whereas DriveVA
jointly denoises video and action latents
\cite{drivelaw2026,driveva2026}.
These approaches differ in how prediction and action are coupled, but their
deployed backbone computation remains predetermined.
\method\ instead determines the required video DiT depth from the quality of
the currently decoded plan.

\paragraph{Efficient video representations and adaptive inference.}
Recent studies distinguish learning predictive video representations from
rendering future pixels.
In embodied robot control, DiT4DiT conditions an action diffuser on
intermediate video-DiT features \cite{dit4dit2026}, while Fast-WAM retains
future-video supervision but bypasses explicit imagination during deployment
\cite{fastwam2026}.
In autonomous driving, DriveLaW similarly uses an early-denoising video
representation for trajectory planning without VAE decoding
\cite{drivelaw2026}.
Together, these studies show that predictive video representations can support
action generation without rendering future observations, but they still rely
on a fixed representation interface or predetermined computation path.
Early-exit networks such as BranchyNet and MSDNet instead attach intermediate
predictors and vary executed depth according to prediction confidence
\cite{branchynet2017,msdnet2018}.
Their criteria target single-label classification and do not directly address
diffusion-based trajectory generation, highly tied planning rewards, or
trajectory-quality verification.
\method\ combines these directions by attaching trajectory exits to a single
video DiT and using predicted planning quality to determine whether additional
backbone blocks are warranted.

\paragraph{Diffusion trajectory planning and planner optimization.}
Diffusion-based planners model multimodal driving actions through iterative
trajectory refinement.
DiffusionDrive truncates trajectory denoising around learned anchors,
Diffusion Planner supports flexible guidance during sampling, and GoalFlow
uses flow matching for goal-conditioned planning
\cite{diffusiondrive2025,zheng2025diffusion,xing2025goalflow}.
ReCogDrive adopts a five-step trajectory diffuser and further improves it
through planner-only DiffGRPO
\cite{recogdrive2025,grpo2024}.
We use the same trajectory-space formulation at every exit, providing a
controlled action-decoding interface for studying intermediate video-DiT
representations and adaptive backbone depth.
After imitation learning, planner-only DiffGRPO refines the exit heads while
leaving the video backbone fixed.
Our quality model is orthogonal to trajectory denoising: rather than guiding
the sampling process, it determines whether the current exit is sufficient or
whether additional world-model computation is required.
Because planning rewards are highly saturated and frequently tied, it predicts
metric components instead of imposing a strict total ordering over candidates.

\section{Motivating Analysis}
\label{sec:motivating_analysis}

On NAVSIM v1, we first use identical fixed-exit, single-trajectory readouts to
test sensitivity to video-noise level and DiT depth, and whether different
depths solve redundant scene sets.
Reported PDMS aggregates validation-best checkpoints over ten seeds.

\begin{table}[t]
\centering
\scriptsize
\begin{tabular}{@{}lrrrrrr@{}}
\toprule
Block & 5 & 9 & 15 & 18 & 22 & 30\\
\midrule
IL & 81.94 & 83.60 & \textbf{86.56} & 84.14 & 83.62 & 80.71\\
Planner RL & 86.02 & 87.56 & \textbf{90.62} & 88.92 & 87.42 & 85.82\\
\bottomrule
\end{tabular}
\captionsetup{skip=2pt}
\caption{Layer-wise planning quality with identical schedules and
validation-best selection over ten seeds.}
\label{tab:layers}
\end{table}

\paragraph{Video timestep is not the main bottleneck.}
At block 15, five sampling indices $\{1,9,17,25,32\}$ produce imitation-learning scores
$\{86.44,86.56,86.57,86.55,86.50\}$, a range of 0.13.
At block 18 the corresponding scores are
$\{84.02,84.14,83.99,84.12,84.01\}$, a range of 0.15.
We therefore fix index 17 and study depth, avoiding an unnecessary search over video-noise levels.

\paragraph{Middle layers are strong, but later layers are not redundant.}
Table~\ref{tab:layers} shows that block 15 is best after both imitation and RL for the single-trajectory readout.
RL improves all six exits by roughly four to five points.

\begin{figure}[!t]
\centering
\includegraphics[width=0.90\columnwidth]{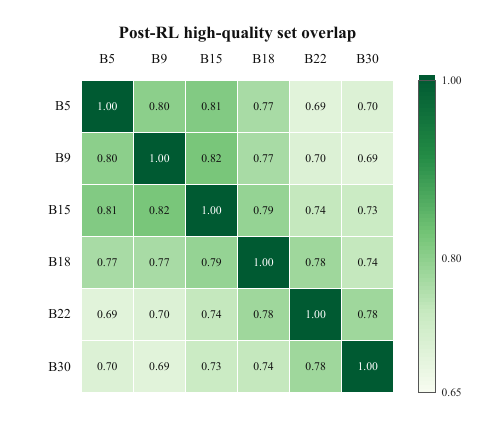}
\caption{Post-RL Jaccard overlap of high-quality scene sets. Off-diagonal
values of 0.69--0.82 show substantial but incomplete sharing across exits.}
\label{fig:jaccard}
\end{figure}

\paragraph{Solved-scene sets remain only partially shared.}
Figure~\ref{fig:jaccard} complements the directional pairwise counts.
The largest off-diagonal overlap is 0.82 for blocks 9 and 15, whereas early--late
pairs fall as low as 0.69.
Thus, global dominance by block 15 does not make all other exits redundant.
The overlap suggests that a planner can terminate when a shallow trajectory is
already strong while retaining deeper computation as a fallback.

\begin{figure*}[!t]
\centering
\includegraphics[width=\textwidth]{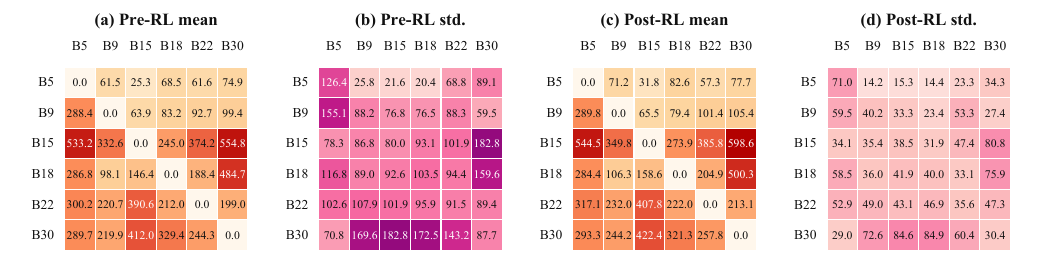}
\caption{Pairwise large-advantage counts before and after planner RL. Panels
report the mean or standard deviation over ten seed-pair runs of
$N_{a\succ b}=N(s_a-s_b\geq50)$.}
\label{fig:pairwise_advantage}
\end{figure*}

\paragraph{Pairwise advantages are structured and reproducible.}
Figure~\ref{fig:pairwise_advantage} exposes information hidden by the global
averages.
Block 15 has the largest directional advantage counts: it outperforms block
30 by at least 50 points on $554.8$ scenes before planner RL and $598.6$
scenes afterward, on average.  The reverse direction remains nonzero
($412.0$ and $422.4$ scenes, respectively), directly showing why a globally
strong fixed exit is not uniformly best scene by scene.
Across all matrix entries, the maximum standard deviation falls from $182.8$
to $84.9$ after planner RL, indicating more stable pairwise counts without
changing the broad depth ordering.
Therefore, no fixed exit dominates every scene; this motivates the per-input,
quality-guided depth allocation introduced next.

\FloatBarrier
\section{Method}

Motivated by Section~\ref{sec:motivating_analysis}, \method\ exposes multiple
trajectory exits and routes by predicted quality.

\subsection{Problem and Backbone}

Let $o=(I,S_{\mathrm{ego}},L_{\mathrm{nav}})$ contain the current front-camera
image, deployment-available current and historical ego states, and navigation
command.
The planner predicts a four-second ego trajectory
$\tau=\{(x_t,y_t,\theta_t)\}_{t=1}^{8}$ at 0.5-second intervals.
Adaptive inference is single-trajectory: every attempted exit contributes one
new trajectory.

We initialize the visual dynamics backbone from Wan2.2-TI2V-5B, a latent video DiT in the Wan family \cite{wan2025,dit2023}.
Let $d(o)$ be the deployment-available text description derived from the
observation.
The conditional Wan branch directly processes the current image--description
pair at fixed video-noise index $s^\star=17$ of the 40-step schedule.
It does not consume a ground-truth or encoded future video at deployment.
For DiT block $\ell$, the hidden representation is
\begin{equation}
    h_\ell = F_{1:\ell}\bigl(I,d(o);s^\star\bigr),
\end{equation}
where $F_{1:\ell}$ denotes the conditional backbone prefix.
This operation is a \emph{single video-feature forward}; it is separate from the five DDIM \cite{ddim2021} steps used by each trajectory head.

\begin{figure*}[!t]
    \centering
    \includegraphics[width=0.93\textwidth]{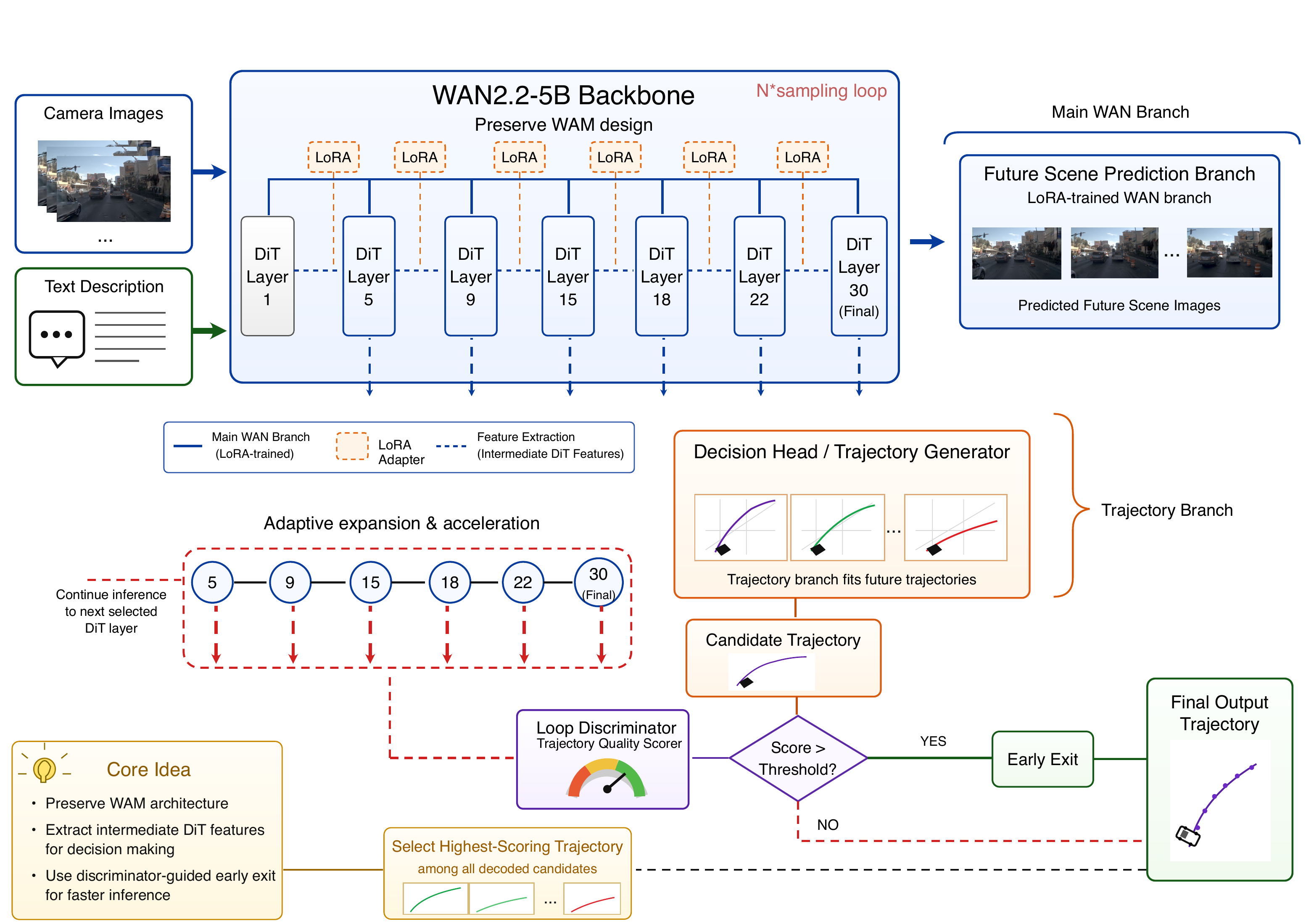}
    \caption{Overview of \method. Wan2.2 retains video supervision while six
    intermediate blocks feed independent ReCogDrive-style trajectory heads.
    At inference, one trajectory is decoded per attempted exit; the lightweight
    DINOv2-Small scorer either returns the best accumulated trajectory or
    continues from the cached hidden state. Repeated heads and LoRA locations
    are schematic, and the future-scene branch denotes training supervision
    rather than the default deployed path.}
    \label{fig:overview}
\end{figure*}

\subsection{Multi-Exit Trajectory Decoding}

We select exits $\mathcal{E}=\{5,9,15,18,22,30\}$.
At exit $\ell$, a projection $P_\ell$ converts $h_\ell$ into
trajectory-conditioning tokens, and an independent diffusion head $G_\ell$
generates one trajectory
\begin{equation}
    \tau_\ell
    =
    G_\ell(P_\ell(h_\ell),S_{\mathrm{ego}},L_{\mathrm{nav}}).
\end{equation}
All heads have the same architecture, optimization budget, batch size, and number of epochs.
They use five action-denoising steps, matching ReCogDrive.
The exit heads have independent parameters and do not exchange features or predictions.
This makes differences across exits attributable to backbone depth rather than to head capacity.

Each single-trajectory head uses the same logged-trajectory diffusion objective
and five-step training protocol as ReCogDrive.

\subsection{Quality-Guided Adaptive Inference}

The scorer fine-tunes a DINOv2-Small image encoder and embeds the flattened
eight-pose trajectory with an MLP.
The image and trajectory features are concatenated and passed to six
independent two-layer MLP heads.
The scorer does not receive ego state or navigation command.
For component set
$\mathcal{R}=\{\mathrm{NC,DAC,DDC,TTC,EP,Comf}\}$, it produces
\begin{equation}
\begin{split}
    \mathbf a_\ell &= S_\phi(I,\tau_\ell),\\
    \hat{\mathbf r}_\ell
    &=\sigma(\mathbf a_\ell)
      =(\widehat{\mathrm{NC}},\widehat{\mathrm{DAC}},
        \widehat{\mathrm{DDC}},\widehat{\mathrm{TTC}},
        \widehat{\mathrm{EP}},\widehat{\mathrm{Comf}}).
\end{split}
\end{equation}
The normalized PDMS composition $\Gamma(\cdot)\in[0,1]$ is expressed on the
reported 100-point scale as
\begin{equation}
    Q(\hat{\mathbf r})=100\,\Gamma(\hat{\mathbf r}).
\end{equation}

Let
\begin{equation}
    \mathcal{A}_j=\{\tau_{\ell_m}\}_{m=1}^{j}
\end{equation}
denote the one trajectory decoded at each of the first $j$ attempted exits;
hence $|\mathcal A_j|=j\leq6$.
We cache the component predictions and maintain
\begin{equation}
\begin{split}
    \hat{\tau}_j
    &=
    \operatorname*{arg\,max}_{\tau_{\ell_m}\in\mathcal{A}_j}
    Q(\hat{\mathbf r}_{\ell_m}),\\
    \hat q_j
    &=
    \max_{m\leq j}Q(\hat{\mathbf r}_{\ell_m}).
\end{split}
\end{equation}
Given $\eta\in[0,100]$, the controller terminates at the first exit satisfying
$\hat q_j\geq\eta$.
If no earlier exit satisfies the threshold, the final exit returns the
highest-scoring trajectory accumulated across all six exits.
Because planning scores are highly saturated and frequently tied, the scorer
predicts metric components and primarily acts as an exit-quality verifier
rather than enforcing a strict total ordering over trajectories.

After a rejected exit, only the previously unevaluated backbone blocks are
executed, while hidden states and trajectory scores are reused.
Let $p_j$ be the probability of terminating at exit $\ell_j$,
$c^{\mathrm{bb}}_{\ell_j}$ the cumulative conditional-backbone cost up to that
exit, and $c^G_{\ell_m}$ and $c^S_{\ell_m}$ the single-trajectory generation
and scoring costs at exit $\ell_m$.
The expected inference cost is
\begin{equation}
\begin{split}
    \mathbb{E}[\mathcal{C}_\eta]
    =
    \sum_{j=1}^{J} p_j
    \biggl[
        c^{\mathrm{bb}}_{\ell_j}
        +\sum_{m=1}^{j}
        \bigl(c^G_{\ell_m}+c^S_{\ell_m}\bigr)
    \biggr].
\end{split}
\end{equation}
The threshold $\eta$ is selected on the validation set to determine the
desired quality--computation operating point.

\begin{algorithm}[t]
\caption{Quality-Guided Layer-Adaptive Planning}
\label{alg:adaptive}
\begin{algorithmic}[1]
\STATE Encode the current image and text description from observation $o$
\STATE Initialize accumulated trajectory pool $\mathcal{A}\leftarrow\emptyset$
\FOR{$\ell \in \{5,9,15,18,22,30\}$}
    \STATE Continue the conditional Wan forward to block $\ell$
    \STATE Decode one trajectory $\tau_\ell$
    \STATE Predict $\hat{\mathbf r}_\ell$ and cache $Q(\hat{\mathbf r}_\ell)$
    \STATE $\mathcal{A}\leftarrow\mathcal{A}\cup\{\tau_\ell\}$
    \STATE Obtain $(\hat q,\hat\tau)$ from the best trajectory in $\mathcal{A}$
    \IF{$\hat q\geq\eta$ \OR $\ell=30$}
        \RETURN $\hat\tau$
    \ENDIF
\ENDFOR
\end{algorithmic}
\end{algorithm}

\subsection{Training}

\paragraph{Video-domain adaptation.}
Each NAVSIM training sample contains the current front-camera frame and the
next eight frames sampled at 2\,Hz, forming a nine-frame clip over four
seconds.
The original $1600{\times}900$ images are resized to the Wan landscape
resolution of $1280{\times}704$.
The text condition is generated from deployment-available structured
attributes, including map metadata, discretized ego speed, an
observed-ego-motion-history-derived maneuver, and traffic density.
The maneuver uses only past and current ego poses and kinematics, never a
future-trajectory label.
Samples without all eight temporally aligned future frames are discarded.
Future frames supervise only the video objective during training; deployment
uses the current frame and does not decode future images.
The caption templates and preprocessing details are provided in
Appendix~\ref{sec:data}, especially Sections~\ref{sec:image_preprocess}
and~\ref{sec:text_condition}.

During imitation learning, the Wan backbone is adapted using LoRA
\cite{lora2022}, while the trajectory projections and heads are optimized in
full.
The actor objective is
\begin{equation}
    \mathcal{L}_{\mathrm{actor}}
    =
    \lambda_{\mathrm{vid}}\mathcal{L}_{\mathrm{vid}}
    +
    \sum_{\ell\in\mathcal{E}}
    \lambda_\ell\mathcal{L}_{\mathrm{traj}}^\ell,
\end{equation}
where $\mathcal{L}_{\mathrm{vid}}$ is the native Wan video-diffusion objective.
Both video supervision and trajectory decoding use the same fixed
video-noise index $s^\star=17$ and the same conditional Wan forward during
joint adaptation; no second backbone forward is required.
Video prediction is retained as representation supervision rather than a
required deployment output.

\paragraph{Trajectory-quality scorer.}
The scorer, including its DINOv2-Small encoder, is fine-tuned on generated
trajectories using evaluator-provided component targets.
All six components use equal-weight soft-label BCE-with-logits:
\begin{equation}
    \mathcal{L}_{\mathrm{score}}
    =
    \sum_i\sum_{m\in\mathcal R}
    \operatorname{BCE}_{\mathrm{logit}}
    \left(a_{i,m},r^{\mathrm{oracle}}_{i,m}\right),
\end{equation}
where each oracle component is used directly as a target in $[0,1]$, without
binarization or component-specific regression losses.
This component-wise soft-target formulation avoids imposing an artificial
ordering among the many tied or near-perfect trajectories.
Generated trajectories are treated as stop-gradient inputs during scorer
training, such that
\begin{equation}
    \nabla_{\theta_{\mathrm{Wan}},\theta_G}
    \mathcal{L}_{\mathrm{score}}=0.
\end{equation}
Thus, the scorer does not alter the actor's trajectory distribution or compete
with the trajectory-generation objective.
The actor and scorer are optimized in alternating updates on the same training
stream, with generated trajectories detached before each scorer update.
With equal branch weight, the complete training-stage objective can be written
for bookkeeping as
\begin{equation}
    \mathcal{L}
    =
    \mathcal{L}_{\mathrm{actor}}
    +
    \mathcal{L}_{\mathrm{score}},
\end{equation}
with gradient isolation between the actor and scorer branches.

\paragraph{Planner-only reinforcement learning.}
After imitation learning, we freeze the Wan backbone and quality scorer and refine each trajectory head using the DiffGRPO \cite{recogdrive2025}.
A complete five-step action-denoising chain is treated as one trajectory
action and receives its NAVSIM evaluator score as reward.
No scorer gradient or routing decision is propagated to the actor during this
stage.
For each random seed, we select the validation-best checkpoint under the same
training protocol; reported layer-wise statistics aggregate ten seeds.
Optimizer settings, LoRA configuration, and loss weights are provided in
Appendix~\ref{sec:training}.

\section{Experiments}

\subsection{Setup}

\paragraph{Datasets and metrics.}
We follow the standard NAVSIM v1 and NAVSIM v2 navtest protocols \cite{navsim2024}.
NAVSIM v1 reports Predictive Driver Model Score (PDMS), combining no-at-fault collision (NC), drivable-area compliance (DAC), time-to-collision (TTC), ego progress (EP), and comfort.
NAVSIM v2 reports the extended EPDMS metric.
Unless noted otherwise, our model uses the front camera without LiDAR and
predicts eight poses over four seconds.
All adaptive-routing and latency results decode one trajectory per attempted
exit.
NAVSIM-to-Wan clips use one anchor plus eight future front-camera frames at
0.5-second intervals; exact retained-sample counts after temporal filtering
are reported in Appendix~\ref{sec:clip_construction}.
For zero-shot evaluation, a model trained on NAVSIM is evaluated on nuScenes without target-domain fine-tuning; we report average L2 displacement and collision rate following prior world-model planners.

\paragraph{Protocols.}
Hardware latency is measured at batch size one on a single A100 80GB.
End-to-end planning latency includes VAE image encoding, the conditional Wan
forward, every attempted trajectory head, and scorer evaluation.
The adaptive planner, fixed block-15 planner, and fixed full-depth planner
average 170, 190, and 320\,ms, respectively.
For context, with cached text context, full 40-step classifier-free
future-video generation takes 13.22\,s under the same hardware setting,
including 12.05\,s of denoising and 1.17\,s of VAE encoding/decoding.

\subsection{Main Results}

\begin{table}[t]
    \centering
    \scriptsize
    \setlength{\tabcolsep}{1.2pt}
    \resizebox{\columnwidth}{!}{%
    \begin{tabular}{@{}lcrrrrrr@{}}
    \toprule
    Method & Input & NC & DAC & TTC & Comf. & EP & PDMS $\uparrow$\\
    \midrule
    \multicolumn{8}{l}{\emph{Traditional end-to-end planners}}\\
    VADv2-$\mathcal{V}_{8192}$ \cite{vadv22024} & C & 97.2 & 89.1 & 91.6 & 100 & 76.0 & 80.9\\
    UniAD \cite{uniad2023} & C & 97.8 & 91.9 & 92.9 & 100 & 78.8 & 83.4\\
    TransFuser \cite{transfuser2022} & CL & 97.7 & 92.8 & 92.8 & 100 & 79.2 & 84.0\\
    PARA-Drive \cite{paradrive2024} & C & 97.9 & 92.4 & 93.0 & 99.8 & 79.3 & 84.0\\
    ReCogDrive-IL \cite{recogdrive2025} & C & 98.1 & 94.7 & 94.2 & 100 & 80.9 & 86.5\\
    DiffusionDrive \cite{diffusiondrive2025} & CL & 98.2 & 96.2 & 94.7 & 100 & 82.2 & 88.1\\
    \midrule
    \multicolumn{8}{l}{\emph{World-model planners}}\\
    LAW \cite{li2024enhancing} & C & 96.4 & 95.4 & 88.7 & 99.9 & 81.7 & 84.6\\
    Epona \cite{epona2025} & C & 97.9 & 95.1 & 93.8 & 99.9 & 80.4 & 86.2\\
    WoTE \cite{wote2025} & CL & 98.5 & 96.8 & 94.9 & 99.9 & 81.9 & 88.3\\
    DriveVLA-W0 \cite{drivevlaw02025} & C & 98.4 & 95.3 & 95.2 & 100 & 80.9 & 87.2\\
    PWM \cite{zhao2025forecasting} & C & 98.6 & 95.9 & 95.4 & 100 & 81.8 & 88.1\\
    DriveVA \cite{driveva2026} & C & 99.2 & 97.5 & 98.7 & 100 & 83.5 & 90.5\\
    \method\ (single trajectory) & C & 98.6 & 97.9 &  95.6 & 100 & 85.1 & 90.8\\
    \method\ (fixed B22, 64 prop.) & C & 99.8 & 98.3 & 98.3 & 100 & 86.6 & \textbf{92.6}\\
    \bottomrule
    \end{tabular}
    }
    \captionsetup{skip=2pt}
    \caption{NAVSIM v1 navtest PDMS. Baselines follow DriveVA; C/CL denote
    camera/camera+LiDAR; our 64-proposal result uses fixed B22 without routing.}
    \label{tab:navsim_v1}
    \end{table}
    
    \begin{table}[t]
    \centering
    \scriptsize
    \setlength{\tabcolsep}{0.7pt}
    \resizebox{\columnwidth}{!}{%
    \begin{tabular}{@{}lcrrrrrrrrrr@{}}
    \toprule
    Method & Input & NC & DAC & DDC & TLC & EP & TTC & LK & HC & EC & EPDMS $\uparrow$\\
    \midrule
    Human Agent & -- & 100 & 100 & 99.8 & 100 & 87.4 & 100 & 100 & 98.1 & 90.1 & 90.3\\
    \midrule
    DiffusionDrive & CL & 98.2 & 95.9 & 99.4 & 99.8 & 87.5 & 97.3 & 96.8 & 98.3 & 87.7 & 84.5\\
    ReCogDrive & C & 98.3 & 95.2 & 99.5 & 99.8 & 87.1 & 97.5 & 96.6 & 98.3 & 86.5 & 83.6\\
    \midrule
    Epona & C & 97.1 & 95.7 & 99.3 & 99.7 & 88.6 & 96.3 & 97.0 & 98.0 & 67.8 & 85.1\\
    DriveVLA-W0 & C & 98.5 & 99.1 & 98.0 & 99.7 & 86.4 & 98.1 & 93.2 & 97.9 & 58.9 & 86.1\\
    \method\  & C & 98.5 & 98.0 & 99.5 & 99.8 & 87.6 & 97.4 & 95.4 & 98.2 & 75.5 & \textbf{89.9}\\
    \bottomrule
    \end{tabular}
    }
    \captionsetup{skip=2pt}
    \caption{NAVSIM v2 navtest EPDMS under the public protocol
    \cite{diffusiondrive2025,recogdrive2025,epona2025,drivevlaw02025};
    C/CL denote camera/camera+LiDAR.}
    \label{tab:navsim_v2}
    \end{table}

Tables~\ref{tab:navsim_v1} and~\ref{tab:navsim_v2} report the two benchmark
versions separately.
On NAVSIM v1, the adaptive single-trajectory planner obtains 90.8 PDMS.
The auxiliary fixed-B22 64-proposal result reaches 92.6 PDMS, exceeding
DriveVA's mixed-data headline result by 1.7 points and its NAVSIM-only result
by 2.1 points; its training details are provided in
Appendix~\ref{sec:fixed64}.
On NAVSIM v2, \method\ reaches 89.9 EPDMS, outperforming the listed front-view
world-model planners.

\paragraph{Zero-shot transfer.}
Without nuScenes fine-tuning, \method\ obtains 0.88\,m average L2 error and
0.08\% collision rate.
DriveVA reports 0.84\,m and 0.06\% under the corresponding zero-shot setting.
However, DriveVA executes the full Wan backbone and generates future images to
support trajectory prediction, making its inference cost substantially higher
and closer to a full Wan video-generation pipeline than to our single
conditional early-exit planning pass.
Table~\ref{tab:zero_shot} retains only averages in the main paper; horizon-wise
results and supervision details are provided in Appendix~\ref{sec:nuscenes}.

\begin{table}[t]
\centering
\scriptsize
\setlength{\tabcolsep}{3pt}
\begin{tabular}{lccc}
\toprule
Method & FT & Avg. L2 $\downarrow$ & Coll. (\%) $\downarrow$\\
\midrule
\multicolumn{4}{l}{\emph{Target-domain tuned}}\\
ST-P3 \cite{stp32022} & \checkmark & 2.11 & 0.71\\
UniAD \cite{uniad2023} & \checkmark & 1.03 & 0.31\\
OccNet \cite{occnet2023} & \checkmark & 2.14 & 0.72\\
OccWorld \cite{occworld2024} & \checkmark & 1.40 & 0.87\\
VAD-Tiny \cite{vad2023} & \checkmark & 1.30 & 0.72\\
VAD-Base \cite{vad2023} & \checkmark & 1.22 & 0.53\\
GenAD \cite{genad2024} & \checkmark & 0.91 & 0.43\\
Doe-1 \cite{doe12024} & \checkmark & 1.26 & 0.53\\
Epona \cite{epona2025} & \checkmark & 1.25 & 0.36\\
\midrule
\multicolumn{4}{l}{\emph{NAVSIM-to-nuScenes zero-shot}}\\
DriveVLA-W0 \cite{drivevlaw02025} & -- & 1.43 & 0.77\\
PWM \cite{zhao2025forecasting} & -- & 3.99 & 0.36\\
DriveVA \cite{driveva2026} & -- & \textbf{0.84} & \textbf{0.06}\\
\method & -- & 0.88 & 0.08\\
\bottomrule
\end{tabular}
\captionsetup{skip=2pt}
\caption{nuScenes comparison using horizon-averaged L2 error and collision
rate; FT denotes target-domain fine-tuning.}
\label{tab:zero_shot}
\end{table}

\subsection{Scorer Reliability and Adaptive Trade-off}

The scorer is evaluated on an offline diagnostic pool covering 12,146 scenes.
Exact top-score selection succeeds in 91.2\% of scenes, and tie-aware soft accuracy reaches 94.4\% when a selected trajectory within 5\% of the true top score is accepted.
Strict rank correlation is not informative here: more than 95\% of scenes
contain trajectory groups that are all perfect, all zero, or tied at the top.
We instead stress-test consequential failures.
Only 51 scenes (0.42\%) select a trajectory at least 50\% worse than an available near-perfect trajectory, and 69 scenes (0.57\%) exceed a 20\% gap.

\begin{table}[t]
\centering
\scriptsize
\begin{tabular}{@{}lccc@{}}
\toprule
Policy & PDMS $\uparrow$ & Exit by B15 & Lat. (ms)\\
\midrule
Fixed B15 & 90.62 & 100\% & 190\\
Adaptive $\eta=70$ & 88.49 & $98.8\%$ & 112\\
Adaptive $\eta=80$ & 90.64 & $95.2\%$ & 143\\
Adaptive $\eta=90$ & \textbf{90.79} & $94.1\%$ & 170\\
Adaptive $\eta=95$ & 90.75 & $65.9\%$ & 284\\
Full path & 85.82 & -- & 320\\
\bottomrule
\end{tabular}
\captionsetup{skip=2pt}
\caption{Adaptive performance--efficiency trade-off on one A100 at batch size
one. Latency is end-to-end; Exit by B15 denotes termination within the first
three exits. The full path includes one VAE decode; $\eta$ uses 100-point $Q$.}
\label{tab:adaptive}
\end{table}

Table~\ref{tab:adaptive} reports the current threshold sweep.
At $\eta=90$, the adaptive policy improves on fixed B15 by 0.17 PDMS while
reducing average end-to-end planning latency by approximately 10\%, from
190 to 170\,ms.
More than 94\% of scenes terminate within the first three exits, and the fixed
full-depth planner averages 320\,ms.
The permissive threshold 70 instead loses 2.13 points relative to fixed B15.

\subsection{Ablations and Efficiency}

\begin{table}[t]
\centering
\scriptsize
\setlength{\tabcolsep}{2.5pt}
\begin{tabular}{lrlr}
\toprule
Adaptation & PDMS & Visual backbone & PDMS\\
\midrule
Wan frozen & 84.20 & ViT-S & 83.91\\
Separate LoRA + cache & 84.95 & ViT-B & 85.62\\
Joint Wan LoRA & 90.62 & ViT-L & 88.88\\
Full Wan tuning & 90.64 & -- & --\\
\bottomrule
\end{tabular}
\captionsetup{skip=2pt}
\caption{Single-trajectory adaptation and visual-backbone ablations on NAVSIM
v1 (PDMS).}
\label{tab:ablation}
\end{table}

\paragraph{Adaptation and visual representation.}
Table~\ref{tab:ablation} shows that freezing Wan is insufficient.
Fine-tuning Wan separately and then caching its features recovers only a small
part of the loss, whereas joint LoRA training improves the single-trajectory
model by 5.67 points over cached features.
Full fine-tuning yields no meaningful gain over LoRA.
Replacing Wan features with ViT-S/B/L lowers single-trajectory PDMS by 6.71/5.00/1.74 points, respectively.

\paragraph{Scorer backbone.}
On the same offline diagnostic trajectory pool, DINO-Small obtains 92.59,
compared with 92.54 for DINO-Base, 91.17/91.20 for ViT-S/B, and
92.19/92.55 for ResNet-34/50.
Wan features offer no meaningful scorer advantage while substantially
increasing online cost: the best Wan exit improves the diagnostic score by
only 0.03 points over DINO-Small.
We therefore fine-tune DINO-Small as the default scorer; the full
scorer-backbone and video-index diagnostics are reported in
Appendix~\ref{sec:scorer}.

\paragraph{Computation.}
A full 40-step classifier-free Wan rollout performs 80 DiT forwards and takes
13.22\,s with cached text context: 12.05\,s for iterative denoising and
1.17\,s for VAE encoding/decoding.
The planning path performs one conditional forward up to the selected exit,
executing a five-step trajectory head and lightweight scorer at each attempted
exit.
On an A100, VAE image encoding costs approximately 50\,ms.
Including the conditional Wan forward, every attempted trajectory head, scorer
evaluation, and VAE image encoding, the selected adaptive planner averages
170\,ms end-to-end.
The fixed block-15 planner averages 190\,ms, while the fixed full-depth planner
averages 320\,ms and additionally performs one VAE decode.
This comparison is kept separate from the 13.22\,s full video-generation
runtime.

\section{Conclusion}

In summary, intermediate video-DiT features are robust planning representations well before full image synthesis completes.
By coupling multi-depth trajectory heads with a tie-aware trajectory-quality controller,
\method\ allocates world-model computation according to the quality of the
current plan.
It achieves strong NAVSIM and zero-shot nuScenes results while exposing a practical path from large generative world models to efficient driving policies.

\bibliography{references}

@article{navsim2024,
  title={Navsim: Data-driven non-reactive autonomous vehicle simulation and benchmarking},
  author={Dauner, Daniel and Hallgarten, Marcel and Li, Tianyu and Weng, Xinshuo and Huang, Zhiyu and Yang, Zetong and Li, Hongyang and Gilitschenski, Igor and Ivanovic, Boris and Pavone, Marco and others},
  journal={Advances in Neural Information Processing Systems},
  volume={37},
  pages={28706--28719},
  year={2024}
}

@inproceedings{nuscenes2020,
  title={nuscenes: A multimodal dataset for autonomous driving},
  author={Caesar, Holger and Bankiti, Varun and Lang, Alex H and Vora, Sourabh and Liong, Venice Erin and Xu, Qiang and Krishnan, Anush and Pan, Yu and Baldan, Giancarlo and Beijbom, Oscar},
  booktitle={Proceedings of the IEEE/CVF conference on computer vision and pattern recognition},
  pages={11621--11631},
  year={2020}
}

@article{wan2025,
  title={Wan: Open and advanced large-scale video generative models},
  author={Wan, Team and Wang, Ang and Ai, Baole and Wen, Bin and Mao, Chaojie and Xie, Chen-Wei and Chen, Di and Yu, Feiwu and Zhao, Haiming and Yang, Jianxiao and others},
  journal={arXiv preprint arXiv:2503.20314},
  year={2025}
}

@inproceedings{dit2023,
  title={Scalable diffusion models with transformers},
  author={Peebles, William and Xie, Saining},
  booktitle={Proceedings of the IEEE/CVF international conference on computer vision},
  pages={4195--4205},
  year={2023}
}

@article{lora2022,
  title={Lora: Low-rank adaptation of large language models.},
  author={Hu, Edward J and Shen, Yelong and Wallis, Phillip and Allen-Zhu, Zeyuan and Li, Yuanzhi and Wang, Shean and Wang, Liang and Chen, Weizhu and others},
  journal={Iclr},
  volume={1},
  number={2},
  pages={3},
  year={2022}
}

@article{ddim2021,
  title={Denoising diffusion implicit models},
  author={Song, Jiaming and Meng, Chenlin and Ermon, Stefano},
  journal={arXiv preprint arXiv:2010.02502},
  year={2020}
}

@article{dinov2,
  title={Dinov2: Learning robust visual features without supervision},
  author={Oquab, Maxime and Darcet, Timoth{\'e}e and Moutakanni, Th{\'e}o and Vo, Huy and Szafraniec, Marc and Khalidov, Vasil and Fernandez, Pierre and Haziza, Daniel and Massa, Francisco and El-Nouby, Alaaeldin and others},
  journal={Transactions on Machine Learning Research Journal},
  year={2024}
}

@inproceedings{diffusiondrive2025,
  title={Diffusiondrive: Truncated diffusion model for end-to-end autonomous driving},
  author={Liao, Bencheng and Chen, Shaoyu and Yin, Haoran and Jiang, Bo and Wang, Cheng and Yan, Sixu and Zhang, Xinbang and Li, Xiangyu and Zhang, Ying and Zhang, Qian and others},
  booktitle={Proceedings of the Computer Vision and Pattern Recognition Conference},
  pages={12037--12047},
  year={2025}
}

@article{recogdrive2025,
  title={Recogdrive: A reinforced cognitive framework for end-to-end autonomous driving},
  author={Li, Yongkang and Xiong, Kaixin and Guo, Xiangyu and Li, Fang and Yan, Sixu and Xu, Gangwei and Zhou, Lijun and Chen, Long and Sun, Haiyang and Wang, Bing and others},
  journal={arXiv preprint arXiv:2506.08052},
  year={2025}
}

@article{drivelaw2026,
  title={Drivelaw: Unifying planning and video generation in a latent driving world},
  author={Xia, Tianze and Li, Yongkang and Zhou, Lijun and Yao, Jingfeng and Xiong, Kaixin and Sun, Haiyang and Wang, Bing and Ma, Kun and Chen, Guang and Ye, Hangjun and others},
  journal={arXiv preprint arXiv:2512.23421},
  year={2025}
}

@article{driveva2026,
  title={Driveva: Video action models are zero-shot drivers},
  author={Liu, Mengmeng and Zhang, Diankun and Liu, Jiuming and Cui, Jianfeng and Xie, Hongwei and Chen, Guang and Ye, Hangjun and Yang, Michael Ying and Nex, Francesco and Cheng, Hao},
  journal={arXiv preprint arXiv:2604.04198},
  year={2026}
}

@article{clover2026,
  title={CLOVER: Closed-loop value estimation and ranking for end-to-end autonomous driving planning},
  author={Ang, Sining and Yang, Yuguang and Chen, Canyu and Wang, Yan},
  journal={arXiv preprint arXiv:2605.15120},
  year={2026}
}

@article{fastwam2026,
  title={Fast-wam: Do world action models need test-time future imagination?},
  author={Yuan, Tianyuan and Dong, Zibin and Liu, Yicheng and Zhao, Hang},
  journal={arXiv preprint arXiv:2603.16666},
  year={2026}
}

@article{dit4dit2026,
  title={Dit4dit: Jointly modeling video dynamics and actions for generalizable robot control},
  author={Ma, Teli and Zheng, Jia and Wang, Zifan and Jiang, Chunli and Cui, Andy and Liang, Junwei and Yang, Shuo},
  journal={arXiv preprint arXiv:2603.10448},
  year={2026}
}

@article{grpo2024,
  title={Deepseekmath: Pushing the limits of mathematical reasoning in open language models},
  author={Shao, Zhihong and Wang, Peiyi and Zhu, Qihao and Xu, Runxin and Song, Junxiao and Bi, Xiao and Zhang, Haowei and Zhang, Mingchuan and Li, YK and Wu, Yang and others},
  journal={arXiv preprint arXiv:2402.03300},
  year={2024}
}

@inproceedings{drivinggpt2025,
  title={Drivinggpt: Unifying driving world modeling and planning with multi-modal autoregressive transformers},
  author={Chen, Yuntao and Wang, Yuqi and Zhang, Zhaoxiang},
  booktitle={Proceedings of the IEEE/CVF International Conference on Computer Vision},
  pages={26890--26900},
  year={2025}
}

@inproceedings{epona2025,
  title={Epona: Autoregressive diffusion world model for autonomous driving},
  author={Zhang, Kaiwen and Tang, Zhenyu and Hu, Xiaotao and Pan, Xingang and Guo, Xiaoyang and Liu, Yuan and Huang, Jingwei and Yuan, Li and Zhang, Qian and Long, Xiao-Xiao and others},
  booktitle={Proceedings of the IEEE/CVF International Conference on Computer Vision},
  pages={27220--27230},
  year={2025}
}

@article{vavim2025,
  title={Vavim and vavam: Autonomous driving through video generative modeling},
  author={Bartoccioni, Florent and Ramzi, Elias and Besnier, Victor and Venkataramanan, Shashanka and Vu, Tuan-Hung and Xu, Yihong and Chambon, Loick and Gidaris, Spyros and Odabas, Serkan and Hurych, David and others},
  journal={arXiv preprint arXiv:2502.15672},
  year={2025}
}

@article{drivevlaw02025,
  title={DriveVLA-W0: World models amplify data scaling law in autonomous driving},
  author={Li, Yingyan and Shang, Shuyao and Liu, Weisong and Zhan, Bing and Wang, Haochen and Wang, Yuqi and Chen, Yuntao and Wang, Xiaoman and An, Yasong and Tang, Chufeng and others},
  journal={arXiv preprint arXiv:2510.12796},
  year={2025}
}

@article{fsdrive2025,
  title={Futuresightdrive: Thinking visually with spatio-temporal cot for autonomous driving},
  author={Zeng, Shuang and Chang, Xinyuan and Xie, Mengwei and Liu, Xinran and Bai, Yifan and Pan, Zheng and Xu, Mu and Wei, Xing},
  journal={Advances in Neural Information Processing Systems},
  volume={38},
  pages={67299--67318},
  year={2026}
}

@article{hu2023gaia,
  title={Gaia-1: A generative world model for autonomous driving},
  author={Hu, Anthony and Russell, Lloyd and Yeo, Hudson and Murez, Zak and Fedoseev, George and Kendall, Alex and Shotton, Jamie and Corrado, Gianluca},
  journal={arXiv preprint arXiv:2309.17080},
  year={2023}
}

@inproceedings{gao2023magicdrive,
  title={Magicdrive: Street view generation with diverse 3d geometry control},
  author={Gao, Ruiyuan and Chen, Kai and Xie, Enze and Hong, Lanqing and Li, Zhenguo and Yeung, Dit-Yan and Xu, Qiang},
  booktitle={International Conference on Learning Representations},
  volume={2024},
  pages={22841--22860},
  year={2024}
}

@inproceedings{wang2024drivedreamer,
  title={Drivedreamer: Towards real-world-drive world models for autonomous driving},
  author={Wang, Xiaofeng and Zhu, Zheng and Huang, Guan and Chen, Xinze and Zhu, Jiagang and Lu, Jiwen},
  booktitle={European conference on computer vision},
  pages={55--72},
  year={2024},
  organization={Springer}
}

@inproceedings{wen2024panacea,
  title={Panacea: Panoramic and controllable video generation for autonomous driving},
  author={Wen, Yuqing and Zhao, Yucheng and Liu, Yingfei and Jia, Fan and Wang, Yanhui and Luo, Chong and Zhang, Chi and Wang, Tiancai and Sun, Xiaoyan and Zhang, Xiangyu},
  booktitle={Proceedings of the IEEE/CVF Conference on Computer Vision and Pattern Recognition},
  pages={6902--6912},
  year={2024}
}

@article{gao2024vista,
  title={Vista: A generalizable driving world model with high fidelity and versatile controllability},
  author={Gao, Shenyuan and Yang, Jiazhi and Chen, Li and Chitta, Kashyap and Qiu, Yihang and Geiger, Andreas and Zhang, Jun and Li, Hongyang},
  journal={Advances in Neural Information Processing Systems},
  volume={37},
  pages={91560--91596},
  year={2024}
}

@inproceedings{li2024drivingdiffusion,
  title={DrivingDiffusion: Layout-guided multi-view driving scenarios video generation with latent diffusion model},
  author={Li, Xiaofan and Zhang, Yifu and Ye, Xiaoqing},
  booktitle={European Conference on Computer Vision},
  pages={469--485},
  year={2024},
  organization={Springer}
}

@article{wang2025mila,
  title={MiLA: Multi-view intensive-fidelity long-term video generation world model for autonomous driving},
  author={Wang, Haiguang and Liu, Daqi and Xie, Hongwei and Liu, Haisong and Ma, Enhui and Yu, Kaicheng and Wang, Limin and Wang, Bing},
  journal={arXiv preprint arXiv:2503.15875},
  year={2025}
}

@article{yang2025resim,
  title={Resim: Reliable world simulation for autonomous driving},
  author={Yang, Jiazhi and Chitta, Kashyap and Gao, Shenyuan and Chen, Long and Shao, Yuqian and Jia, Xiaosong and Li, Hongyang and Geiger, Andreas and Yue, Xiangyu and Chen, Li},
  journal={Advances in Neural Information Processing Systems},
  volume={38},
  pages={167710--167741},
  year={2026}
}

@article{li2025omninwm,
  title={Omninwm: Omniscient driving navigation world models},
  author={Li, Bohan and Ma, Zhuang and Du, Dalong and Peng, Baorui and Liang, Zhujin and Liu, Zhenqiang and Guo, Xianda and Zhu, Zheng and Ma, Chao and Jin, Yueming and others},
  journal={arXiv preprint arXiv:2510.18313},
  year={2025}
}

@inproceedings{li2024enhancing,
  title={Enhancing end-to-end autonomous driving with latent world model},
  author={Li, Yingyan and Fan, Lue and He, Jiawei and Wang, Yuqi and Chen, Yuntao and Zhang, Zhaoxiang and Tan, Tieniu},
  booktitle={International Conference on Learning Representations},
  volume={2025},
  pages={42942--42959},
  year={2025}
}

@article{hu2024drivingworld,
  title={DrivingWorld: Constructing world model for autonomous driving via video GPT},
  author={Hu, Xiaotao and Yin, Wei and Jia, Mingkai and Deng, Junyuan and Guo, Xiaoyang and Zhang, Qian and Long, Xiaoxiao and Tan, Ping},
  journal={arXiv preprint arXiv:2412.19505},
  year={2024}
}

@inproceedings{min2024driveworld,
  title={Driveworld: 4d pre-trained scene understanding via world models for autonomous driving},
  author={Min, Chen and Zhao, Dawei and Xiao, Liang and Zhao, Jian and Xu, Xinli and Zhu, Zheng and Jin, Lei and Li, Jianshu and Guo, Yulan and Xing, Junliang and others},
  booktitle={Proceedings of the IEEE/CVF conference on computer vision and pattern recognition},
  pages={15522--15533},
  year={2024}
}

@article{zhao2025forecasting,
  title={From forecasting to planning: Policy world model for collaborative state-action prediction},
  author={Zhao, Zhida and Fu, Talas and Wang, Yifan and Wang, Lijun and Lu, Huchuan},
  journal={Advances in Neural Information Processing Systems},
  volume={38},
  pages={134585--134611},
  year={2026}
}

@inproceedings{zheng2025diffusion,
  title={Diffusion-based planning for autonomous driving with flexible guidance},
  author={Zheng, Yinan and Liang, Ruiming and Zheng, Kexin and Zheng, Jinliang and Mao, Liyuan and Li, Jianxiong and Gu, Weihao and Ai, Rui and Li, Shengbo and Zhan, Xianyuan and others},
  booktitle={International conference on learning representations},
  volume={2025},
  pages={37207--37227},
  year={2025}
}

@inproceedings{xing2025goalflow,
  title={Goalflow: Goal-driven flow matching for multimodal trajectories generation in end-to-end autonomous driving},
  author={Xing, Zebin and Zhang, Xingyu and Hu, Yang and Jiang, Bo and He, Tong and Zhang, Qian and Long, Xiaoxiao and Yin, Wei},
  booktitle={Proceedings of the Computer Vision and Pattern Recognition Conference},
  pages={1602--1611},
  year={2025}
}

@article{vadv22024,
  title={Vadv2: End-to-end vectorized autonomous driving via probabilistic planning},
  author={Chen, Shaoyu and Jiang, Bo and Gao, Hao and Liao, Bencheng and Xu, Qing and Zhang, Qian and Huang, Chang and Liu, Wenyu and Wang, Xinggang},
  journal={arXiv preprint arXiv:2402.13243},
  year={2024}
}

@inproceedings{uniad2023,
  title={Planning-oriented autonomous driving},
  author={Hu, Yihan and Yang, Jiazhi and Chen, Li and Li, Keyu and Sima, Chonghao and Zhu, Xizhou and Chai, Siqi and Du, Senyao and Lin, Tianwei and Wang, Wenhai and others},
  booktitle={Proceedings of the IEEE/CVF conference on computer vision and pattern recognition},
  pages={17853--17862},
  year={2023}
}

@article{transfuser2022,
  title={Transfuser: Imitation with transformer-based sensor fusion for autonomous driving},
  author={Chitta, Kashyap and Prakash, Aditya and Jaeger, Bernhard and Yu, Zehao and Renz, Katrin and Geiger, Andreas},
  journal={IEEE transactions on pattern analysis and machine intelligence},
  volume={45},
  number={11},
  pages={12878--12895},
  year={2022},
  publisher={IEEE}
}

@inproceedings{paradrive2024,
  title={Para-drive: Parallelized architecture for real-time autonomous driving},
  author={Weng, Xinshuo and Ivanovic, Boris and Wang, Yan and Wang, Yue and Pavone, Marco},
  booktitle={Proceedings of the IEEE/CVF Conference on Computer Vision and Pattern Recognition},
  pages={15449--15458},
  year={2024}
}

@inproceedings{wote2025,
  title={End-to-end driving with online trajectory evaluation via bev world model},
  author={Li, Yingyan and Wang, Yuqi and Liu, Yang and He, Jiawei and Fan, Lue and Zhang, Zhaoxiang},
  booktitle={Proceedings of the IEEE/CVF International Conference on Computer Vision},
  pages={27137--27146},
  year={2025}
}

@inproceedings{stp32022,
  title={St-p3: End-to-end vision-based autonomous driving via spatial-temporal feature learning},
  author={Hu, Shengchao and Chen, Li and Wu, Penghao and Li, Hongyang and Yan, Junchi and Tao, Dacheng},
  booktitle={European Conference on Computer Vision},
  pages={533--549},
  year={2022},
  organization={Springer}
}

@inproceedings{occnet2023,
  title={Scene as occupancy},
  author={Tong, Wenwen and Sima, Chonghao and Wang, Tai and Chen, Li and Wu, Silei and Deng, Hanming and Gu, Yi and Lu, Lewei and Luo, Ping and Lin, Dahua and others},
  booktitle={Proceedings of the IEEE/CVF International Conference on Computer Vision},
  pages={8406--8415},
  year={2023}
}

@inproceedings{occworld2024,
  title={Occworld: Learning a 3d occupancy world model for autonomous driving},
  author={Zheng, Wenzhao and Chen, Weiliang and Huang, Yuanhui and Zhang, Borui and Duan, Yueqi and Lu, Jiwen},
  booktitle={European conference on computer vision},
  pages={55--72},
  year={2024},
  organization={Springer}
}

@inproceedings{vad2023,
  title={Vad: Vectorized scene representation for efficient autonomous driving},
  author={Jiang, Bo and Chen, Shaoyu and Xu, Qing and Liao, Bencheng and Chen, Jiajie and Zhou, Helong and Zhang, Qian and Liu, Wenyu and Huang, Chang and Wang, Xinggang},
  booktitle={Proceedings of the IEEE/CVF International Conference on Computer Vision},
  pages={8340--8350},
  year={2023}
}

@inproceedings{genad2024,
  title={Genad: Generative end-to-end autonomous driving},
  author={Zheng, Wenzhao and Song, Ruiqi and Guo, Xianda and Zhang, Chenming and Chen, Long},
  booktitle={European Conference on Computer Vision},
  pages={87--104},
  year={2024},
  organization={Springer}
}

@article{doe12024,
  title={Doe-1: Closed-loop autonomous driving with large world model},
  author={Zheng, Wenzhao and Xia, Zetian and Huang, Yuanhui and Zuo, Sicheng and Zhou, Jie and Lu, Jiwen},
  journal={arXiv preprint arXiv:2412.09627},
  year={2024}
}

@article{branchynet2017,
  title={Branchynet: Fast inference via early exiting from deep neural networks},
  author={Teerapittayanon, Surat and McDanel, Bradley and Kung, Hsiang-Tsung},
  journal={arXiv preprint arXiv:1709.01686},
  year={2017}
}

@inproceedings{msdnet2018,
  title={Multi-scale dense networks for resource efficient image classification},
  author={Huang, Gao and Chen, Danlu and Li, Tianhong and Wu, Felix and Van Der Maaten, Laurens and Weinberger, Kilian},
  booktitle={International conference on learning representations},
  year={2018}
}

\clearpage
\appendix
\setcounter{secnumdepth}{2}
\section{Appendix Roadmap}

This appendix provides the details deferred from the main paper: NAVSIM
evaluation and data construction, caption generation and image preprocessing,
training and gradient isolation, the auxiliary fixed-exit 64-proposal model,
extended video-noise and DiT-depth analyses, scorer diagnostics, adaptive
routing, latency decomposition, and horizon-level nuScenes comparisons.
Table~\ref{tab:coverage} maps each explicit pointer in the main paper to its
corresponding appendix section.

\begin{table}[H]
\centering
\scriptsize
\setlength{\tabcolsep}{3pt}
\begin{tabular}{p{0.42\columnwidth}p{0.49\columnwidth}}
\toprule
Main-paper pointer & Appendix location\\
\midrule
Caption templates and image preprocessing &
Section~\ref{sec:data}, especially Secs.~\ref{sec:image_preprocess}
and~\ref{sec:text_condition}\\
Optimizer, LoRA, loss, and gradient-isolation details &
Section~\ref{sec:training}\\
NAVSIM-to-Wan temporal filtering and corpus accounting &
Section~\ref{sec:clip_construction}\\
Fixed-B22 64-proposal training &
Section~\ref{sec:fixed64}\\
Scorer backbone and video-index diagnostics &
Section~\ref{sec:scorer}\\
nuScenes horizon-wise comparison and supervision status &
Section~\ref{sec:nuscenes}\\
Adaptive routing and latency scope &
Sections~\ref{sec:routing} and~\ref{sec:latency}\\
\bottomrule
\end{tabular}
\caption{Coverage of details explicitly deferred from the main paper to the
appendix.}
\label{tab:coverage}
\end{table}

\section{Datasets, Metrics, and Evaluation}
\label{sec:evaluation}

\subsection{NAVSIM Protocol}

\paragraph{Input and output.}
The default agent uses the latest \texttt{CAM\_F0} image, ego state, and
route-level navigation command; it does not use LiDAR.
The planner predicts eight ego poses
$\tau=\{(x_t,y_t,\theta_t)\}_{t=1}^{8}$ at 0.5-second intervals in the
ego-centric rear-axle frame, giving a four-second horizon.
The route command is supplied by the benchmark and does not encode obstacle
or traffic-light state.

\paragraph{NAVSIM v1.}
We use the official navtest evaluation protocol \cite{navsim2024}.
NAVSIM v1 evaluates each submitted trajectory through a four-second
non-reactive simulation: background actors follow their recorded futures,
while a controller rolls out the ego vehicle along the submitted plan.
Its Predictive Driver Model Score (PDMS) combines multiplicative safety and
feasibility penalties with a weighted measure of progress, time-to-collision,
and comfort:
\begin{equation}
\mathrm{PDMS}
=\mathrm{NC}\cdot\mathrm{DAC}\cdot
\frac{5\,\mathrm{EP}+5\,\mathrm{TTC}+2\,\mathrm{Comf}}{12}.
\label{eq:pdms_v1}
\end{equation}
Here NC denotes no-at-fault collision and DAC denotes drivable-area
compliance. We use $Q=100\,\mathrm{PDMS}$ when expressing scorer thresholds
on the 100-point scale used by the paper's tables.
Multiplicative safety terms and saturated sub-scores create many exact ties
among otherwise different trajectories.

\paragraph{NAVSIM v2.}
NAVSIM v2 extends this protocol with driving-direction compliance (DDC),
traffic-light compliance (TLC), lane keeping (LK), history comfort (HC), and
extended comfort (EC). NC and DDC take values in
$\{0,\tfrac{1}{2},1\}$; DAC and TLC are binary multipliers. EP is continuous
in $[0,1]$, while TTC, LK, HC, and EC are binary. The additive weights are 5
for EP and TTC and 2 for LK, HC, and EC.

NAVSIM v2 also filters false-positive penalties against the human agent. For
a component $m$, let
\begin{equation}
F_m(a,h)=
\begin{cases}
1, & m(h)=0,\\
m(a), & \text{otherwise},
\end{cases}
\label{eq:human_filter}
\end{equation}
where $a$ and $h$ denote the submitted agent and the human reference.
Thus, a violation that is also triggered by the human rollout is neutralized
rather than attributed solely to the submitted planner. With
$\mathcal M=\{\mathrm{NC,DAC,DDC,TLC}\}$ and
$\mathcal W=\{\mathrm{TTC,EP,HC,LK,EC}\}$, the per-stage score is
\begin{equation}
\mathrm{EPDMS}=
\left(\prod_{m\in\mathcal M}F_m(a,h)\right)
\frac{\sum_{m\in\mathcal W}w_mF_m(a,h)}
{\sum_{m\in\mathcal W}w_m}.
\label{eq:epdms_v2}
\end{equation}

To approximate closed-loop behavior without interactive simulation, v2 uses
a two-stage aggregation. It first scores the initial four-second scene, then
scores precomputed follow-up scenes that begin from alternative end states.
Follow-up scores are aggregated with a Gaussian kernel according to the
distance between each follow-up start state and the submitted planner's
first-stage end state. The final result multiplies the first-stage score by
this weighted second-stage score. The reported 89.9 EPDMS uses the official
evaluator and a single front camera without LiDAR.

\paragraph{Split discipline.}
NAVSIM training uses navtrain-derived records.
The v1 and v2 test splits are used only for evaluation; no privileged map,
future occupancy, or future sensor observation is supplied to the deployed
planner.
Privileged evaluator state is used only for offline metric computation and,
for the auxiliary 64-proposal model, pseudo-expert construction.

\subsection{nuScenes Zero-Shot Protocol}

The NAVSIM-trained model is evaluated on nuScenes \cite{nuscenes2020} without
nuScenes fine-tuning.
The model continues to use a single front view and predicts ego trajectories
in the local coordinate frame.
We follow prior world-model planners by reporting displacement error and
collision rate at 1, 2, and 3 seconds and their horizon average.
Section~\ref{sec:nuscenes} separates methods trained on nuScenes from
NAVSIM-to-nuScenes zero-shot methods.

\section{NAVSIM-to-Wan Training Data}
\label{sec:data}

\subsection{Clip Construction and Temporal Filtering}
\label{sec:clip_construction}

For an anchor at time $t_0$, it retrieves the current \texttt{CAM\_F0} frame
and the next eight front-camera frames at the native NAVSIM/OpenScene rate of
2\,Hz:
\begin{equation}
    \mathcal V =
    \{I(t_0),I(t_0+0.5),\ldots,I(t_0+4.0)\}.
\end{equation}
Timestamps and each frame's relative time are stored in the metadata and
checked when constructing the clip.
An anchor is removed if any of the eight required future frames is absent,
which commonly occurs near the end of a log, or if a referenced camera image
cannot be loaded.
The aligned trajectory target contains the eight future
$(x,y,\theta)$ poses over the same four seconds.

Starting from 103,288 candidate navtrain tokens, temporal filtering retains
82,555 complete nine-frame clips and removes 20,733 candidates that cannot
provide the required current-plus-eight-future sequence. NAVSIM does not
define an official train/validation partition for this converted corpus, so
we report corpus-level counts rather than presenting an internal model-
selection partition as an official benchmark split.

\begin{table}[H]
\centering
\small
\begin{tabular}{ll}
\toprule
Field & Setting\\
\midrule
Camera & \texttt{CAM\_F0}\\
Source image & $1600{\times}900$\\
Wan image & $1280{\times}704$\\
Video target & 1 anchor + 8 future frames\\
Frame interval & 0.5\,s (2\,Hz)\\
Planning horizon & 4.0\,s\\
Trajectory target & $8{\times}(x,y,\theta)$\\
Storage & Metadata plus relative image paths\\
\bottomrule
\end{tabular}
\caption{Construction of one NAVSIM-to-Wan training item. The nine-frame clip
satisfies Wan2.2-TI2V-5B's $4n+1$ frame-count constraint.}
\label{tab:data_item}
\end{table}

The metadata record stores the scenario token, log identifier, anchor
timestamp, relative camera paths, camera intrinsics, ego velocity and
acceleration, aligned future trajectory, and programmatic text description.
The PyTorch dataset resolves image paths relative to the OpenScene root at
load time.

\subsection{Image Preprocessing}
\label{sec:image_preprocess}

The source $1600{\times}900$ RGB image is converted to Wan's landscape
resolution of $1280{\times}704$ before video-domain adaptation.
The Wan branch receives RGB values mapped to the normalization expected by
the frozen Wan VAE.
The scorer has a separate image path: the same current front image is resized
to $832{\times}480$ for the DINOv2 encoder and normalized with ImageNet
statistics.
Only the current image is used online; future images are training targets for
the video objective and are never provided to the deployed planner.
No image-captioning network, detector output, map raster, or future frame is
used to create the deployed visual input.

\subsection{Programmatic Text Descriptions}
\label{sec:text_condition}

The Wan text condition is generated from structured, auditable attributes
rather than from a separate vision-language captioner. The final template is
\begin{quote}
\small
\texttt{Vehicle \{motion\}\{turn\_info\} at \{speed:.1f\} m/s in urban
environment.}
\end{quote}
The motion descriptor uses the current speed: below 0.5\,m/s is
\emph{stationary}, 0.5--3\,m/s is \emph{slowly}, 3--8\,m/s is
\emph{cruising}, and above 8\,m/s is \emph{fast}. A turn descriptor is added
only when the observed historical lateral displacement exceeds 0.3\,m.
Current map-area and traffic-flow metadata are also incorporated by the
description generator. Crucially, all fields are computed from map metadata
and observations at or before the anchor time; neither the ground-truth
future trajectory nor future images are queried. The prompt is stored with
the scenario token, and the finite set of resulting T5 text embeddings is
cached in both training and inference.

\section{Architecture and Training Details}
\label{sec:training}

\subsection{Fixed Video-Noise Index and Multi-Exit Readout}

Wan2.2-TI2V-5B \cite{wan2025} contains 30 video-DiT blocks.
The native video sampler has 40 noise steps, but the deployed planner executes
one conditional Wan forward at fixed sampling index $s^\star=17$.
The future part of the latent is initialized from scheduler noise at that
index; the current image provides the conditioned latent slice.
No observed future latent, unconditional classifier-free-guidance branch, or
video VAE decoding is used by the planning path.

Hidden states are read after blocks
\begin{equation}
    \mathcal E=\{5,9,15,18,22,30\}.
\end{equation}
Each readout has its own projection and trajectory head.
The six heads have independent parameters and do not exchange features or
predictions, while their trajectory losses all backpropagate through the
shared Wan LoRA during imitation learning.
When routing continues from one exit to the next, the previously evaluated
DiT prefix and its hidden state are reused.

\subsection{Single-Trajectory Diffusion Head}

The adaptive system uses one ReCogDrive-style action DiT at every exit
\cite{recogdrive2025}.
Each head starts from a Gaussian trajectory latent and performs five DDIM
updates \cite{ddim2021}, conditioned on projected video-DiT features, the
current ego state, and the navigation command.
It decodes one eight-pose trajectory per attempted exit.
Consequently, the cumulative pool contains at most six trajectories, one from
each evaluated depth; it never contains $6\times64$ proposals.

\subsection{Actor, Scorer, and Gradient Isolation}

During imitation learning, the Wan backbone is adapted with LoRA
\cite{lora2022}; the trajectory projections and heads are fully trainable.
The actor loss is
\begin{equation}
    \mathcal L_{\mathrm{actor}}
    =
    \lambda_{\mathrm{vid}}\mathcal L_{\mathrm{vid}}
    +
    \sum_{\ell\in\mathcal E}
    \lambda_\ell\mathcal L_{\mathrm{traj}}^\ell .
\end{equation}
The native video-diffusion objective and all trajectory heads use the same
fixed video-noise index and conditional Wan forward, avoiding a second
backbone pass.

The quality scorer receives detached trajectories.
Thus,
\begin{equation}
    \nabla_{\theta_{\mathrm{Wan}},\theta_G}
    \mathcal L_{\mathrm{score}}=0,
\end{equation}
and scorer fitting cannot reshape the trajectory generator through proposal
coordinates.
Actor and scorer updates alternate on the same training stream.

After imitation learning, Wan and the scorer are frozen.
Only the six trajectory heads are refined using the DiffGRPO formulation
adopted by ReCogDrive: one complete five-step trajectory-denoising chain is
treated as a composite action and receives the NAVSIM planning score as its
reward.
Each exit uses the same training budget and validation-best checkpoint rule.
Layer statistics reported in the main paper aggregate ten runs.

\subsection{LoRA Placement and Reproducibility Manifest}

LoRA modules are applied throughout the Wan backbone, rather than only at the
six readout blocks shown schematically in the main architecture figure.
The base Wan parameters and VAE remain frozen; projection layers, trajectory
heads, and LoRA parameters are trainable during joint imitation learning.
Table~\ref{tab:manifest} records the settings established by the experiment
records.

\begin{table*}[t]
\centering
\small
\begin{tabular}{p{0.23\textwidth}p{0.29\textwidth}p{0.39\textwidth}}
\toprule
Component & Setting & Scope\\
\midrule
Backbone & Wan2.2-TI2V-5B & Conditional branch; 30 video-DiT blocks\\
Video schedule & 40 sampling steps; planning index 17 & One conditional
forward for planning and joint supervision\\
Exit blocks & 5, 9, 15, 18, 22, 30 & Independent projections and heads\\
Trajectory output & 8 poses at 0.5\,s & Four-second horizon\\
Trajectory sampler & Five DDIM updates & Same ReCogDrive-style protocol at
all exits\\
Adaptive proposals & One per attempted exit & At most six accumulated\\
Scorer & DINOv2-Small, fine-tuned at $832{\times}480$ & Current front camera plus candidate
trajectory; no ego state or command\\
Checkpoint selection & Validation-best & Same rule for all exits\\
Planner RL & Wan and scorer frozen & Trajectory heads only\\
Randomness & Ten runs & Used for reported layer-wise analyses\\
\bottomrule
\end{tabular}
\caption{Architecture and training configuration for the adaptive
single-trajectory system.}
\label{tab:manifest}
\end{table*}

The single-trajectory and auxiliary 64-proposal systems use the same Wan
LoRA recipe; their planning decoders differ. The shared backbone-adaptation
configuration is listed in Table~\ref{tab:lora_hparams}. Non-LoRA Wan
parameters and the VAE are frozen.

\begin{table}[H]
\centering
\small
\begin{tabular}{ll}
\toprule
Wan LoRA item & Setting\\
\midrule
Target modules & $q,k,v,o$\\
Rank / alpha / dropout & 32 / 64 / 0.05\\
Optimizer & AdamW\\
Base learning rate & $2\times10^{-4}$\\
Per-GPU batch / GPUs & 5 / 4\\
Gradient accumulation & 4\\
Effective batch & 80\\
Training length & 80 epochs\\
Precision / strategy & bf16-mixed / DDP\\
Planning index / frames & 17 / 9\\
Wan input / feature dim. & $1280{\times}704$ / 3072\\
\bottomrule
\end{tabular}
\caption{Shared Wan-LoRA backbone-adaptation configuration.}
\label{tab:lora_hparams}
\end{table}

Table~\ref{tab:single_train} gives the ReCogDrive-style action-head schedule.
The imitation stage trains the small action DiT and its $3072\!\to\!384$
feature projection with a 100-step diffusion training schedule and five DDIM
steps at inference. The planner-only GRPO stage initializes both the current
and frozen reference policies from the validation-best imitation checkpoint.

\begin{table*}[t]
\centering
\scriptsize
\setlength{\tabcolsep}{4pt}
\begin{tabular}{lcc}
\toprule
Item & Imitation learning & Planner-only GRPO\\
\midrule
Optimizer / LR & AdamW, $10^{-4}$ & AdamW, $10^{-4}$\\
Betas / weight decay & $(0.9,0.95)$ / $10^{-4}$ & $(0.9,0.95)$ / $10^{-4}$\\
Schedule & 3-epoch warmup + cosine to $10^{-6}$ & cosine, no warmup, minimum 0\\
Training length & 200 epochs & 50 epochs\\
Per-GPU batch / GPUs & 256 / 4 & 8 / 4\\
Effective global batch & 1024 & 32\\
Precision / grad. clip & 16-mixed / 1.0 & 16-mixed / 1.0\\
DDP strategy & standard DDP & find-unused-parameters DDP\\
Trainable planner module & feature projection + action DiT & action DiT only\\
Checkpoint rule & lowest validation loss; retain top 5 & lowest validation loss; retain top 5\\
\bottomrule
\end{tabular}
\caption{Single-trajectory planner optimization.}
\label{tab:single_train}
\end{table*}

For GRPO, each scene forms a group of eight sampled trajectories. Advantages
are standardized within the group, denoising-step contributions are
discounted by $\gamma=0.6$, and log probabilities are clamped to $[-5,2]$.
The loss adds a behavior-cloning term with weight 0.1. Reward weights are
10/5/2/0 for progress/TTC/comfort/driving direction, with a four-second
proposal horizon and 0.1-second simulator interval. Wan and the scorer remain
frozen throughout this stage.

\section{Auxiliary Fixed-B22 64-Proposal Model}
\label{sec:fixed64}

The 92.6 PDMS result is an auxiliary fixed-exit comparator.
It reads block 22 once, generates $K=64$ trajectories at that exit, scores the
64 trajectories, and returns the highest-scoring one.
It does not use the adaptive controller, visit six exits, or combine proposal
sets across depths.

\paragraph{Generator.}
The fixed-exit model uses one learned token for each proposal.
The current ego state is embedded into a 256-dimensional token and added to
the proposal embeddings.
A four-block trajectory decoder applies proposal self-attention, cross-attends
to projected Wan scene tokens, and predicts an eight-pose
$(x,y,\theta)$ trajectory from each token.
The hidden dimension is 256, the feed-forward dimension is 1024, and each
refinement block uses one attention head. The decoder uses 16 projected scene
tokens, four generator refinements, four scorer refinements, and a two-pose
long-horizon auxiliary target. This branch uses the shared LoRA configuration
in Table~\ref{tab:lora_hparams}: four GPUs, batch size 5 per GPU, four-step
gradient accumulation, 80 epochs, and bf16 mixed precision.

\paragraph{Logged and pseudo-expert coverage.}
The target construction follows the evaluator-filtered pseudo-expert protocol
of CLOVER \cite{clover2026}.
Candidate families vary target speed, lateral offset, transition length,
acceleration/deceleration, stop--go behavior, approach braking, and off-road
boundary cases.
Candidates first undergo inexpensive validity checks and are then scored with
the true NAVSIM evaluator using training-time map and future occupancy.
The retained target set covers both the logged trajectory and multiple
high-quality alternatives.

\begin{table}[H]
\centering
\small
\begin{tabular}{ll}
\toprule
Pseudo-expert item & Setting\\
\midrule
Future poses & 8 at 0.5\,s\\
Speed candidates (m/s) & 0, 2, 4, 6, 8, 10, 12, 15\\
Regular lateral offsets (m) & $-3.5$ to $3.5$\\
Boundary offsets (m) & $\{-7.0,-5.5,5.5,7.0\}$\\
Acceleration rates & $\{-2,-1,-0.5,0.5,1,2\}$\\
Maximum scored per scene & 180\\
Retained per scene & 50\\
Coverage top-$K$ & 8\\
Score threshold & 0.8\\
Coverage-loss weight & 0.5\\
\bottomrule
\end{tabular}
\caption{CLOVER-derived pseudo-expert target-construction settings used by
the auxiliary proposal-coverage training.}
\label{tab:pseudo_expert}
\end{table}

For every refinement output, the logged-trajectory term selects the closest
of the 64 generated trajectories under mean posewise L1 distance.
The pseudo-expert coverage term performs the reverse set assignment: for each
retained pseudo expert, it selects the closest generated trajectory.
The current configuration applies the pseudo-expert term to the final
refinement output, uses the top eight pseudo experts above score 0.8, and
weights this term by 0.5.
The scorer uses evaluator-provided component labels on detached generated
trajectories.
The active logged-trajectory and final-score losses each have unit weight.
This auxiliary training should not be confused with the five-step
single-trajectory diffusion objective used by adaptive routing.

\section{Extended Video-Noise Analysis}
\label{sec:timestep}

\begin{table}[H]
\centering
\small
\setlength{\tabcolsep}{3.5pt}
\begin{tabular}{lccccc}
\toprule
Video index & 1 & 9 & 17 & 25 & 32\\
\midrule
B15, single & 86.44 & 86.56 & 86.57 & 86.55 & 86.50\\
B18, single & 84.02 & 84.14 & 83.99 & 84.12 & 84.01\\
\midrule
Fixed B15, 64 prop. & 92.01 & 92.12 & 92.11 & 92.05 & 92.07\\
Fixed B18, 64 prop. & 92.45 & 92.55 & 92.59 & 92.45 & 92.43\\
\bottomrule
\end{tabular}
\caption{Planning scores at five tested video sampling indices. The first two
rows are single-trajectory imitation diagnostics; the final two rows use a
fixed-exit 64-proposal model and no adaptive routing.}
\label{tab:timestep}
\end{table}

Across the five tested indices, the range is 0.13 PDMS at block 15 and
0.15 at block 18 for the single-trajectory planner.
The 64-proposal diagnostic shows similarly small ranges of 0.11 and 0.14.
These results support robustness to the tested noise levels; they do not claim
invariance to every possible video timestep or scheduler.
Index 17 is fixed for all subsequent analysis in the paper.

\section{Extended DiT-Depth Analysis}
\label{sec:depth}

\subsection{Fixed-Exit Planning Scores}

\begin{table}[H]
\centering
\small
\begin{tabular}{lrrrrrr}
\toprule
Block & 5 & 9 & 15 & 18 & 22 & 30\\
\midrule
IL & 81.94 & 83.60 & \textbf{86.56} & 84.14 & 83.62 & 80.71\\
RL & 86.02 & 87.56 & \textbf{90.62} & 88.92 & 87.42 & 85.82\\
Gain & 4.08 & 3.96 & 4.06 & 4.78 & 3.80 & 5.11\\
\bottomrule
\end{tabular}
\caption{Layer-wise single-trajectory PDMS before and after planner-only RL.
All exits use the same schedule and validation-best selection rule.}
\label{tab:layer_scores}
\end{table}

Block 15 is the strongest fixed exit in both stages, but later blocks retain
scene-specific advantages.
This distinction motivates routing by candidate quality rather than always
choosing either block 15 or the final block.

\subsection{Post-RL High-Quality Scene Overlap}

For run $r$ and exit $\ell$, define the high-quality set
\begin{equation}
    \mathcal H_\ell^{(r)}
    =
    \{o:Q_\ell^{(r)}(o)\geq90\}.
\end{equation}
Table~\ref{tab:jaccard} reports the mean intersection-over-union of these sets
across ten aligned runs.

\begin{table}[H]
\centering
\scriptsize
\setlength{\tabcolsep}{3pt}
\begin{tabular}{lrrrrrr}
\toprule
Jaccard & B5 & B9 & B15 & B18 & B22 & B30\\
\midrule
B5  & 1.00 & 0.80 & 0.81 & 0.77 & 0.69 & 0.70\\
B9  & 0.80 & 1.00 & 0.82 & 0.77 & 0.70 & 0.69\\
B15 & 0.81 & 0.82 & 1.00 & 0.79 & 0.74 & 0.73\\
B18 & 0.77 & 0.77 & 0.79 & 1.00 & 0.78 & 0.74\\
B22 & 0.69 & 0.70 & 0.74 & 0.78 & 1.00 & 0.78\\
B30 & 0.70 & 0.69 & 0.73 & 0.74 & 0.78 & 1.00\\
\bottomrule
\end{tabular}
\caption{Post-RL Jaccard overlap of scene sets with PDMS at least 90.}
\label{tab:jaccard}
\end{table}

Off-diagonal overlaps range from 0.69 to 0.82.
Thus, the exits solve substantially overlapping but non-identical scene sets:
an early exit is sufficient for most scenes, while deeper exits can still
recover cases not solved by the globally strongest intermediate block.

\subsection{Qualitative Layer-wise Trajectories}

\begin{figure*}[t]
\centering
\begin{minipage}[t]{0.32\textwidth}
\centering
\includegraphics[width=\linewidth]{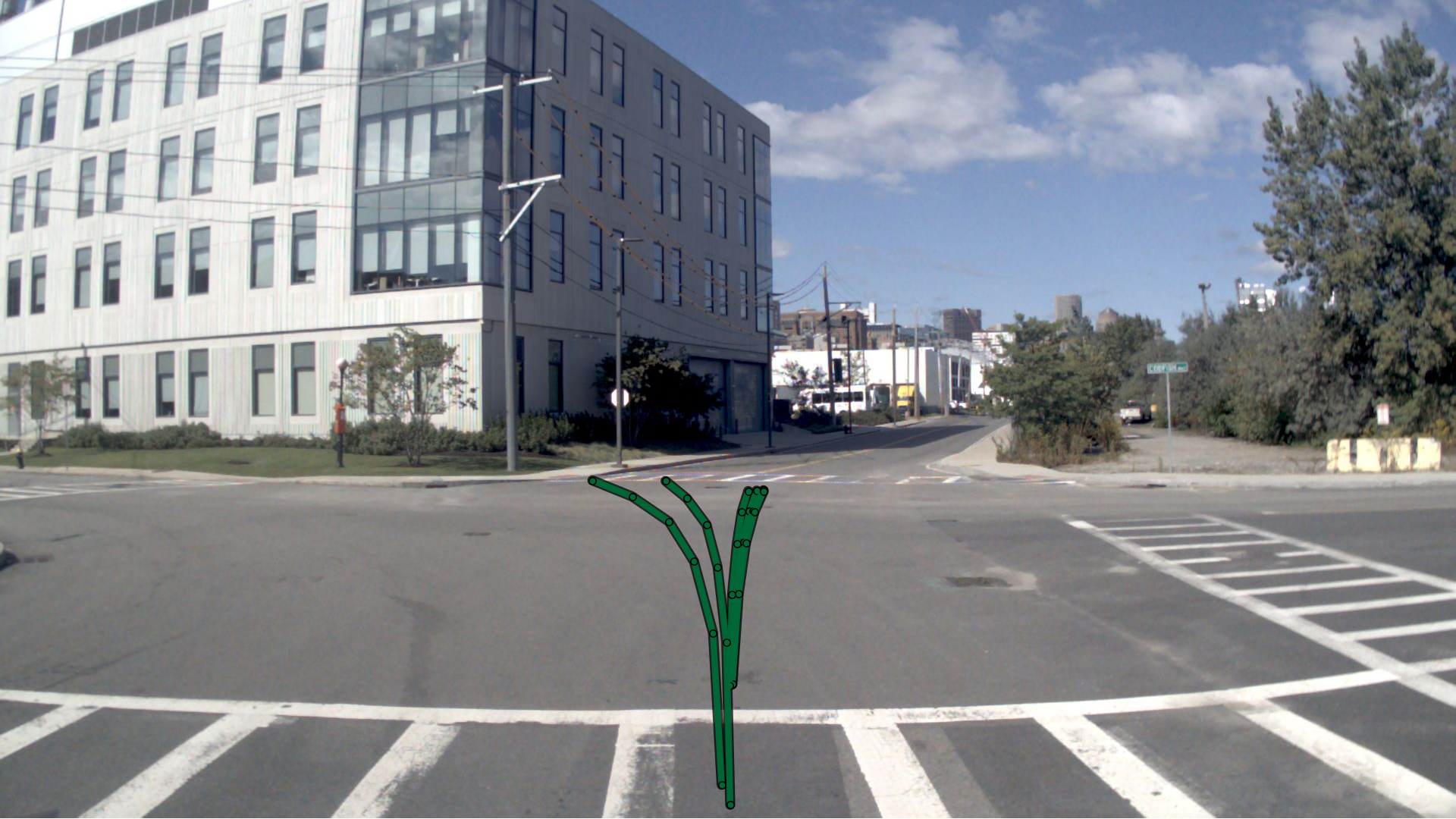}\\[-2pt]
\small (a) Early exits differ from the deeper consensus.
\end{minipage}\hfill
\begin{minipage}[t]{0.32\textwidth}
\centering
\includegraphics[width=\linewidth]{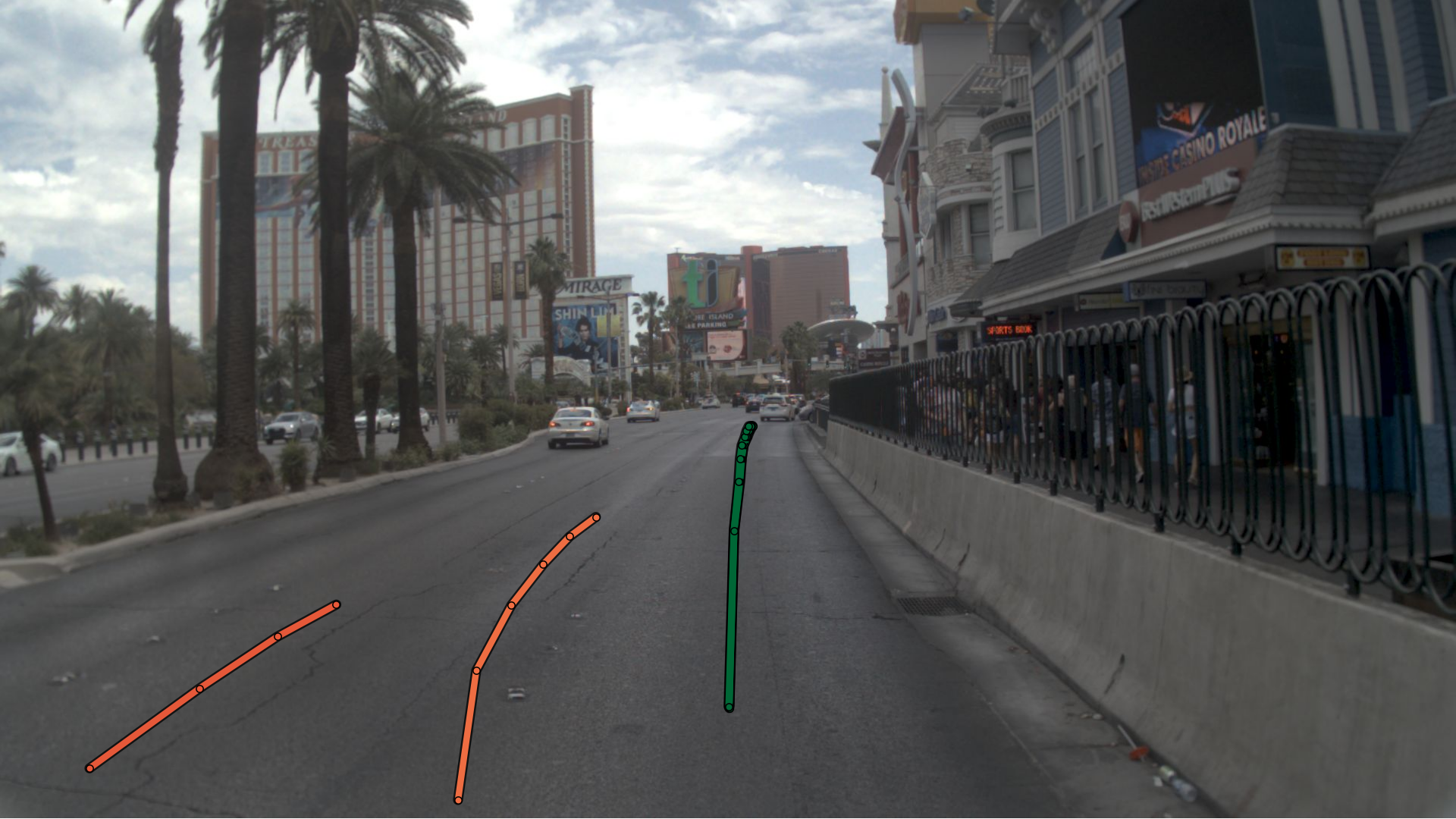}\\[-2pt]
\small (b) Four exits overlap while two retain distinct modes.
\end{minipage}\hfill
\begin{minipage}[t]{0.32\textwidth}
\centering
\includegraphics[width=\linewidth]{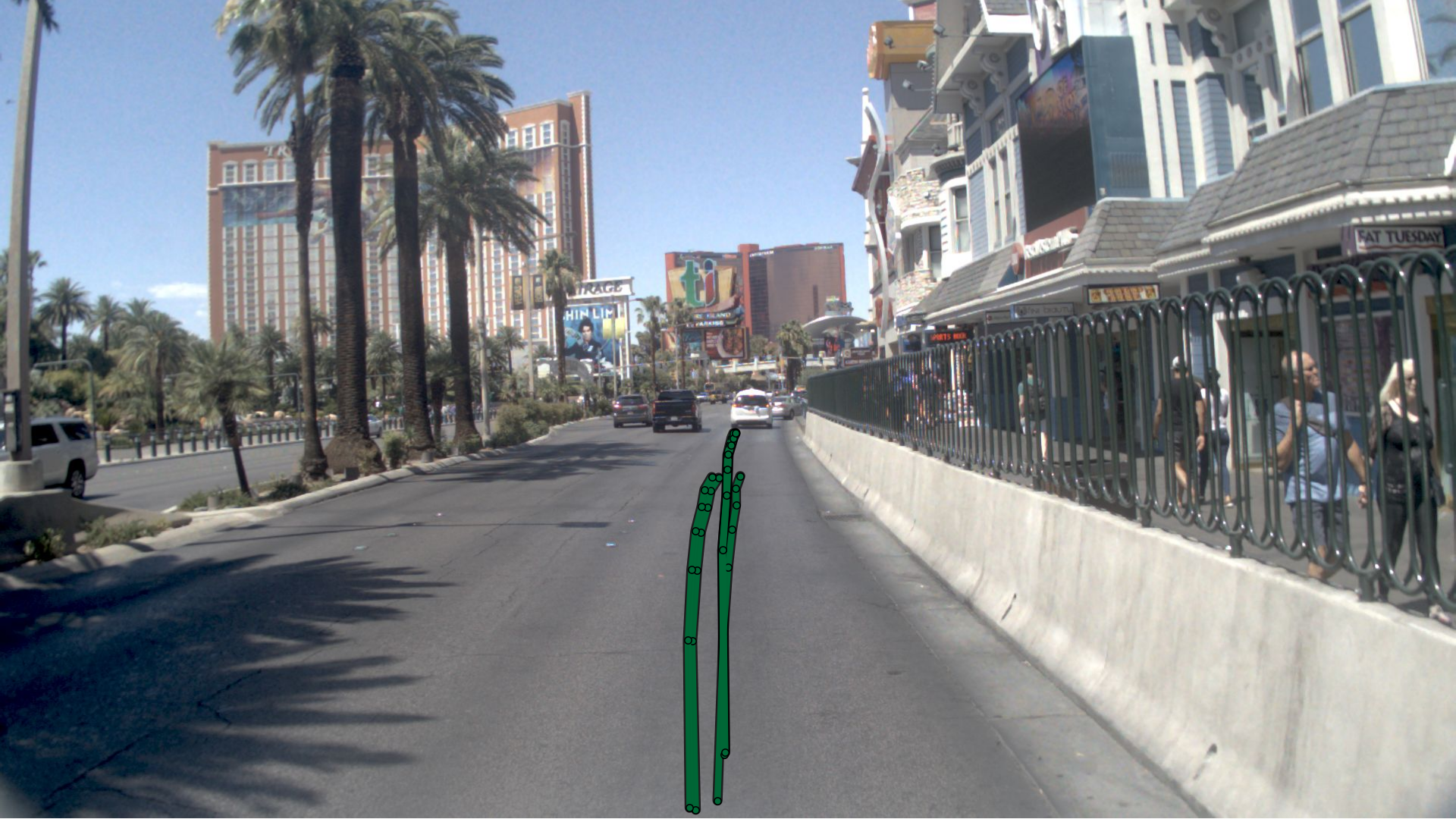}\\[-2pt]
\small (c) All six trajectories receive full score despite variation.
\end{minipage}
\caption{Selected layer-wise trajectory overlays. Trajectory colors encode
scores: green marks high-scoring trajectories, whereas red marks erroneous
trajectories. These cases illustrate both cross-depth
complementarity and the score ties that motivate a verifier rather than a
strict total-order ranker.}
\label{fig:qualitative_depth}
\end{figure*}

Figure~\ref{fig:qualitative_depth} visualizes three recurring patterns behind
the aggregate overlap statistics. The examples are diagnostic overlays of
the six fixed-exit trajectories, not additional model inputs. They show that
layer-wise differences can correspond to distinct decisions, while metric
saturation can also assign identical high scores to visibly different but
acceptable plans.

\section{Trajectory-Quality Scorer}
\label{sec:scorer}

\subsection{Architecture and Objective}

The deployed scorer fine-tunes a DINOv2-Small image encoder
\cite{dinov2}.
It uses only the current front image and the candidate trajectory; ego state
and navigation command are not scorer inputs.
The $8{\times}3$ trajectory is flattened and embedded by an MLP.
The trajectory and image features are concatenated, and six independent
two-layer MLP heads output logits for NC, DAC, DDC, TTC, EP, and Comf.

For evaluator component targets $r^{\mathrm{oracle}}_{i,c}\in[0,1]$, the
scorer uses equal-weight soft-label BCE:
\begin{equation}
\begin{split}
    \mathcal L_{\mathrm{score}}
    =
    \sum_i
    \sum_{c\in\mathcal R}
    \operatorname{BCE}_{\mathrm{logit}}
    (a_{i,c},r^{\mathrm{oracle}}_{i,c}),\\
    \mathcal R
    =
    \{\mathrm{NC,DAC,DDC,TTC,EP,Comf}\}.
\end{split}
\end{equation}
Targets are not binarized.
No global rank loss is used because the official composition contains many
ties and near-ties.
At inference, sigmoid component predictions are combined by
$Q=100\,\Gamma$.

\subsection{Tie-Aware Reliability}

The scorer diagnostic uses a fixed offline candidate pool covering 12,146
scenes.
This pool is used only to compare scorers and is distinct from adaptive
inference, which accumulates at most six trajectories.
More than 95\% of diagnostic scenes contain candidates that are jointly
perfect, jointly zero, or tied at the top.
We therefore report selection quality and consequential errors rather than a
strict total-order correlation.

\begin{table}[H]
\centering
\small
\begin{tabular}{lr}
\toprule
Diagnostic & Rate\\
\midrule
Exact top-score selection & 91.2\%\\
Selection within 5 points & 94.4\%\\
Failure with $\geq50$-point gap & 0.42\%\\
Failure with $\geq20$-point gap & 0.57\%\\
\bottomrule
\end{tabular}
\caption{Tie-aware scorer diagnostics on 12,146 scenes.}
\label{tab:scorer_reliability}
\end{table}

\subsection{Video-Index and Backbone Diagnostics}

\begin{table}[H]
\centering
\small
\setlength{\tabcolsep}{3.5pt}
\begin{tabular}{lccccc}
\toprule
Wan exit & Index 1 & 9 & 17 & 25 & 32\\
\midrule
B15 & 92.01 & 92.12 & 92.11 & 92.05 & 92.07\\
B18 & 92.45 & 92.55 & \textbf{92.59} & 92.45 & 92.43\\
\bottomrule
\end{tabular}
\caption{Wan-based scorer pretest at five video sampling indices. Entries are
true selected-trajectory scores on the same cached candidate pool.}
\label{tab:scorer_timestep}
\end{table}

\begin{table}[H]
\centering
\small
\begin{tabular}{lc}
\toprule
Scorer backbone & Selected-trajectory score\\
\midrule
Wan-B5 & 92.24\\
Wan-B9 & 92.44\\
Wan-B15 & 92.11\\
Wan-B18 & 92.59\\
Wan-B22 & \textbf{92.62}\\
Wan-B30 & 92.57\\
\midrule
DINO-Small & \textbf{92.59}\\
DINO-Base & 92.54\\
ViT-Small & 91.17\\
ViT-Base & 91.20\\
ResNet-34 & 92.19\\
ResNet-50 & 92.55\\
\bottomrule
\end{tabular}
\caption{Scorer-backbone diagnostic on a fixed candidate set. These values
measure the true score of the selected candidate and are not end-to-end
planner PDMS.}
\label{tab:scorer_backbone}
\end{table}

Wan scorer features are also insensitive to the five tested video indices.
At index 17, the best Wan exit obtains 92.62, only 0.03 above DINO-Small.
Because a Wan-based scorer would add a large world-model forward at every
attempted exit, this negligible diagnostic difference does not justify its
online cost.
DINO-Small is therefore used as the quality verifier.

\section{Additional Ablations}
\label{sec:ablations}

\begin{table}[H]
\centering
\scriptsize
\setlength{\tabcolsep}{3pt}
\begin{tabular}{lcc}
\toprule
Wan training & Single & Fixed B22, 64 prop.\\
\midrule
Frozen & 84.20 & 89.91\\
Separate LoRA + cached features & 84.95 & 90.80\\
Joint LoRA & \textbf{90.62} & \textbf{92.59}\\
Full fine-tuning & 90.64 & 92.54\\
\bottomrule
\end{tabular}
\caption{Effect of Wan adaptation. The 64-proposal column is a fixed-exit
auxiliary model without adaptive routing.}
\label{tab:wan_ablation}
\end{table}

Joint training is important: relative to training the head on separately
cached features, joint LoRA improves the single-trajectory model by 5.67
points.
Full Wan fine-tuning provides only 0.02 additional points, so LoRA is used for
the main model.

\begin{table}[H]
\centering
\small
\begin{tabular}{lcc}
\toprule
Visual backbone & Single & Fixed exit, 64 prop.\\
\midrule
ViT-Small & 83.91 & 92.17\\
ViT-Base & 85.62 & 92.21\\
ViT-Large & 88.88 & 92.31\\
Wan intermediate features & \textbf{90.62} & \textbf{92.59}\\
\bottomrule
\end{tabular}
\caption{Static visual features versus intermediate video-DiT features.}
\label{tab:visual_ablation}
\end{table}

The single-trajectory comparison most directly exposes representation
quality: Wan improves over ViT-Large by 1.74 points and over ViT-Small by
6.71.
The gap narrows in the fixed 64-proposal setting because the scorer can choose
among many candidates, but Wan features remain best.

\section{Adaptive Routing}
\label{sec:routing}

At each exit, one trajectory is decoded, scored, and added to the cumulative
pool.
The controller returns the highest-scoring accumulated trajectory as soon as
its predicted quality exceeds threshold $\eta$.
If no threshold is met, block 30 returns the highest-scoring trajectory among
all six candidates.

\begin{table}[H]
\centering
\small
\begin{tabular}{lrrr}
\toprule
Policy & PDMS $\uparrow$ & Exit by B15 & Latency (ms)\\
\midrule
Fixed B15 & 90.62 & 100.0\% & 190\\
Adaptive $\eta=70$ & 88.49 & 98.8\% & 112\\
Adaptive $\eta=80$ & 90.64 & 95.2\% & 143\\
Adaptive $\eta=90$ & \textbf{90.79} & 94.1\% & 170\\
Adaptive $\eta=95$ & 90.75 & 65.9\% & 284\\
Full path & 85.82 & 0.0\% & 320\\
\bottomrule
\end{tabular}
\caption{Adaptive single-trajectory threshold sweep. ``Exit by B15'' denotes
termination at blocks 5, 9, or 15. Latency is batch-one end-to-end planning
time on one A100 80GB.}
\label{tab:appendix_adaptive}
\end{table}

The selected threshold is $\eta=90$.
It improves on the strongest fixed single-trajectory exit by 0.17 PDMS while
reducing mean latency from 190 to 170\,ms.
The threshold sweep also shows why routing cannot be replaced by an
unconditional shallow exit: the permissive $\eta=70$ policy is fastest but
loses 2.13 points relative to fixed B15.

\section{Latency Scope and Profiling}
\label{sec:latency}

\paragraph{Planning latency.}
End-to-end planning latency includes current-image VAE encoding, the
conditional Wan prefix evaluated by the routing decision, every attempted
five-step trajectory head, and every scorer evaluation.
It excludes text encoding because text context is cached.
The measured batch-one means on a single NVIDIA A100 80GB are 190\,ms for fixed
B15, 170\,ms for adaptive routing at $\eta=90$, and 320\,ms for the fixed
full-depth path.
The full-depth reference evaluates all 30 Wan blocks and includes one VAE
decode.

\paragraph{Full video generation.}
Full Wan synthesis follows a different computation graph: 40 denoising steps,
a conditional and unconditional Wan forward at every step, classifier-free
guidance, scheduler updates, and video VAE decoding.
The profiling run uses one anchor plus eight future frames, giving nine
frames, latent shape $[48,3,30,52]$, and sequence length 1170. The prompt
vocabulary is finite and fixed, so its T5 embeddings are cached for both
training and inference. The deployment-comparable video path therefore
includes denoising and both VAE operations, but not online T5 encoding or
file saving.

\begin{table}[H]
\centering
\small
\begin{tabular}{lr}
\toprule
Full-video component & Time\\
\midrule
40-step denoising loop & 12.05\,s\\
Mean conditional DiT per step & 149.40\,ms\\
Mean unconditional DiT per step & 147.80\,ms\\
VAE image encoding & 0.27\,s\\
VAE video decoding & 0.90\,s\\
Cached-text video path, no saving & 13.22\,s\\
T5 encoding in uncached profiling call & 8.34\,s\\
Video saving & 2.87\,s\\
Complete raw profiling call & 21.64\,s\\
Peak allocated memory & 31.19\,GiB\\
\bottomrule
\end{tabular}
\caption{Measured decomposition of one nine-frame Wan generation run. The
13.22\,s deployment-comparable total is
$12.05+0.27+0.90$\,s. Component timers in the raw profiling call are reported
as instrumented and need not sum because stages and bookkeeping overlap.}
\label{tab:video_latency}
\end{table}

The primary 320-to-170\,ms claim compares two planning paths and is therefore
kept separate from full video generation.
The large synthesis cost arises from 80 Wan DiT forwards
($40$ steps $\times$ conditional/unconditional branches), whereas the planner
uses one conditional pass only up to the selected depth and never decodes a
future video.

\section{Detailed nuScenes Comparison}
\label{sec:nuscenes}

\begin{table}[H]
\centering
\scriptsize
\setlength{\tabcolsep}{3pt}
\begin{tabular}{lrrrr}
\toprule
Method & \multicolumn{4}{c}{L2 error (m) $\downarrow$}\\
\cmidrule(lr){2-5}
 & 1\,s & 2\,s & 3\,s & Avg.\\
\midrule
\multicolumn{5}{l}{\emph{Target-domain tuned}}\\
ST-P3 \cite{stp32022} & 1.33 & 2.11 & 2.90 & 2.11\\
UniAD \cite{uniad2023} & 0.48 & 0.96 & 1.65 & 1.03\\
OccNet \cite{occnet2023} & 1.29 & 2.13 & 2.99 & 2.14\\
OccWorld \cite{occworld2024} & 0.52 & 1.27 & 2.41 & 1.40\\
VAD-Tiny \cite{vad2023} & 0.60 & 1.23 & 2.06 & 1.30\\
VAD-Base \cite{vad2023} & 0.54 & 1.15 & 1.98 & 1.22\\
GenAD \cite{genad2024} & 0.36 & 0.83 & 1.55 & 0.91\\
Doe-1 \cite{doe12024} & 0.50 & 1.18 & 2.11 & 1.26\\
Epona \cite{epona2025} & 0.61 & 1.17 & 1.98 & 1.25\\
\midrule
\multicolumn{5}{l}{\emph{NAVSIM-to-nuScenes zero-shot}}\\
DriveVLA-W0 \cite{drivevlaw02025} & 0.43 & 1.26 & 2.60 & 1.43\\
PWM \cite{zhao2025forecasting} & 2.06 & 3.91 & 6.00 & 3.99\\
DriveVA \cite{driveva2026} & 0.33 & 0.76 & 1.43 & \textbf{0.84}\\
\method & 0.35 & 0.71 & 1.58 & 0.88\\
\bottomrule
\end{tabular}
\caption{Horizon-level nuScenes L2 comparison. Published baseline values
follow the DriveVA protocol.}
\label{tab:nuscenes_detailed}
\end{table}

\begin{table}[H]
\centering
\scriptsize
\setlength{\tabcolsep}{3pt}
\begin{tabular}{lrrrr}
\toprule
Method & \multicolumn{4}{c}{Collision rate (\%) $\downarrow$}\\
\cmidrule(lr){2-5}
 & 1\,s & 2\,s & 3\,s & Avg.\\
\midrule
\multicolumn{5}{l}{\emph{Target-domain tuned}}\\
ST-P3 \cite{stp32022} & 0.23 & 0.62 & 1.27 & 0.71\\
UniAD \cite{uniad2023} & 0.05 & 0.17 & 0.71 & 0.31\\
OccNet \cite{occnet2023} & 0.21 & 0.59 & 1.37 & 0.72\\
OccWorld \cite{occworld2024} & 0.12 & 0.40 & 2.08 & 0.87\\
VAD-Tiny \cite{vad2023} & 0.31 & 0.53 & 1.33 & 0.72\\
VAD-Base \cite{vad2023} & 0.04 & 0.39 & 1.17 & 0.53\\
GenAD \cite{genad2024} & 0.06 & 0.23 & 1.00 & 0.43\\
Doe-1 \cite{doe12024} & 0.04 & 0.37 & 1.19 & 0.53\\
Epona \cite{epona2025} & 0.01 & 0.22 & 0.85 & 0.36\\
\midrule
\multicolumn{5}{l}{\emph{NAVSIM-to-nuScenes zero-shot}}\\
DriveVLA-W0 \cite{drivevlaw02025} & 0.22 & 0.66 & 1.42 & 0.77\\
PWM \cite{zhao2025forecasting} & 0.12 & 0.15 & 0.86 & 0.36\\
DriveVA \cite{driveva2026} & 0.00 & 0.07 & 0.12 & \textbf{0.06}\\
\method & 0.00 & 0.09 & 0.15 & 0.08\\
\bottomrule
\end{tabular}
\caption{Horizon-level nuScenes collision-rate comparison.}
\label{tab:nuscenes_collision}
\end{table}

The zero-shot group is trained on NAVSIM and evaluated without nuScenes
fine-tuning.
\method\ obtains 0.88\,m average L2 error and 0.08\% average collision rate
using a single front camera.
DriveVA is slightly better on these two averages, but its inference executes
the full Wan backbone and generates future images to support planning,
whereas \method\ performs a single conditional early-exit planning pass.

\section{Failure Analysis and Interpretation}

The scorer diagnostic contains 51 large failures at a 50-point gap and 69 at
a 20-point gap among 12,146 scenes.
These events are rare but important because the scorer determines whether
additional world-model computation is needed.
They should be interpreted together with the threshold sweep: a stricter
threshold decreases early exits but cannot guarantee safety.
\method\ remains a learned offline planner, and passing the quality threshold
is not a formal safety certificate.

\end{document}